\documentclass[12pt]{spieman}  
\usepackage{amsmath,amsfonts,amssymb}
\usepackage{graphicx}
\usepackage{setspace}
\usepackage{tocloft}
\usepackage{hyperref}
\usepackage{booktabs}
\usepackage{algorithmic}
\usepackage{textcomp}
\usepackage{color}
\makeatletter
\let\c@lofdepth\relax
\let\c@lotdepth\relax
\makeatother
\usepackage{subfigure}
\usepackage{multirow}
\usepackage{rotating}
\usepackage{diagbox}
\usepackage{stfloats}

\title{Gastric histopathology image segmentation using a hierarchical
conditional random field \footnote{This paper is published in Biocybernetics and Biomedical Engineering (BBE), 2020, 40 (4): 1535-1555. Link: https://www.sciencedirect.com/science/article/pii/S0208521620301170.}}

\author[a,b]{Changhao Sun }
\author[a,b,*]{Chen Li}
\author[a,b]{Jinghua Zhang}
\author[a,b]{Muhammad Rahaman}
\author[a,b]{Shiliang Ai}
\author[a,b,e]{Hao Chen}
\author[a,b]{Frank Kulwa}
\author[a,b]{Yixin Li}
\author[c]{Xiaoyan Li}
\author[d]{Tao Jiang}
\affil[a]{Northeastern University, Microscopic Image and Medical Image Analysis Group, 
MBIE College,  Shenyang, P.R. China, 110819
}
\affil[b]{Northeastern University, Engineering Research Center of Medical Imaging and Intelligent Analysis, 
Ministry of Education, Shenyang, P.R. China, 110819}
\affil[c]{China Medical University, Department of Pathology, Cancer Hospital, Liaoning Cancer Hospital and Institute,Shenyang, P.R. China, 110042}
\affil[d]{Chengdu University of Information Technology, Control Engineering College,  
Chengdu, P.R. China, 610103}
\affil[e]{Nanjing University of Science and Technology, School of Computer Science and Engineering, Nanjing, P.R. China, 210094}

\cftpagenumbersoff{figure}
\cftpagenumbersoff{table} 
\begin{document}
\maketitle

\begin{abstract}
For the Convolutional Neural Networks (CNNs) applied in the intelligent diagnosis 
of gastric cancer, existing methods mostly focus on individual characteristics or network frameworks without 
a policy to depict the integral information. Mainly, Conditional Random Field (CRF), 
an efficient and stable algorithm for analyzing images containing complicated contents, can characterize spatial relation in images.
In this paper, a novel Hierarchical Conditional Random Field (HCRF) based Gastric Histopathology Image Segmentation (GHIS) method is proposed, 
which can automatically localize abnormal (cancer) regions in gastric histopathology 
images obtained by an optical microscope to assist histopathologists in medical work. 
This HCRF model is built up with higher order potentials, including pixel-level and 
patch-level potentials, and graph-based post-processing is applied to further improve its segmentation performance. Especially, a CNN is trained to build up the pixel-level potentials 
and another three CNNs are fine-tuned to build up the patch-level 
potentials for sufficient spatial segmentation information. In the experiment, a hematoxylin and eosin (H$\&$E) stained gastric histopathological dataset with 560 abnormal images are divided into training, validation and test sets with a ratio of $1:1:2$. Finally, segmentation 
accuracy, recall and specificity of $78.91\%$, $65.59\%$, and $81.33\%$ are achieved on the test set. 
Our HCRF model demonstrates high segmentation performance and shows its 
effectiveness and future potential in the GHIS field. 
\end{abstract}

\keywords{Image Segmentation, Gastric Cancer, Histopathology Image,  Conditional Random Field, 
Convolutional Neural Network, Feature Extraction}

{\noindent \footnotesize\textbf{*}Chen Li,  \linkable{lichen201096@hotmail.com} }

\begin{spacing}{2}   


\section{Introduction}
\label{s:int} 
Gastric cancer is one of the five most frequent sorts of malignant 
tumors in human beings based on the World Health Organization (WHO) report~\cite{Stewart-2014-WCR}. 
Patients with gastric cancer accounts for $7\%$ of all cancer cases and $9\%$ of 
cancer death cases.
Almost $75\%$ of new cases occur in Asia, and more than $40\%$ occur 
in China~\cite{Stewart-2014-WCR}. The disease specific survival is practically 12 
months and $90\%$ of gastric cancer cases die within the first five years. 
Since gastric cancer is one of the most aggressive and deadliest cancer, it is 
very important for medical professionals to accurately estimate patient 
prognoses~\cite{Garcia-2017-ALD}.

Recently, image analysis systems have found great use in the intelligent diagnosis 
of gastric cancer, where a variety of deep learning (DL) 
methods, especially Convolutional Neural Networks (CNNs), are developed 
and applied to Gastric Histopathology Image Segmentation (GHIS) and 
classification tasks~\cite{Srinidhi-2019-DNN}. From VGG-16~\cite{Simonyan-2014-VDC} to fully convolutional 
networks (FCNs)~\cite{Long-2015-FCN}, from FCNs to DeepLab~\cite{Chen-2018-DSI}, the 
DL algorithms are continually in progress in the GHIS field. Comparing to computer vision for natural images,
lack of training data along with accurate annotations in
the medical image field has currently become a primary problem. To this end, many researches have evidenced
that transfer learning using fine-tuning techniques for CNNs can boost the performance and alleviate the
lack of training data in some degree. The strategy is transferring
a general CNN pretrained by large-scale image
datasets (such as ImageNet~\cite{Russakovsky-2015-ILS}), to a more specific one corresponding
to more complicated tasks~\cite{Shin-2016-DCN, Yosinski-2014-HTR, Qu-2018-GPI}.

There exists two kinds of tasks in GHIS which are 
cell-scale and tissue-scale segmentation~\cite{Srinidhi-2019-DNN, Qu-2018-GPI}. In cell-scale 
segmentation~\cite{Sharma-2015-AMS, Wienert-2012-DSC, Kumar-2017-ADA, Cui-2018-ADL}, researchers 
concentrate on the nuclei information. In tissue-scale 
segmentation~\cite{Peng-2018-FCN, Li-2018-GNA,  AGC-2019-Sun,  Wang-2019-RRM,  WSB-2019-Liang}, all abnormal areas are of interest instead of the separated cells. Nevertheless, the current methods routinely concentrate on individual characteristics, such as loss function and activation function, or 
network frameworks, like as layer numbers and network structures. Lacking the policy
to depict the integral information, which are global properties such as intensity, color and texture. Hence, for achieving a higher computing performance, 
some superior algorithms are designed to incorporate these independent existing approaches.

Particularly, Conditional Random Field (CRF), is an efficient and stable algorithm for 
analyzing the images containing complicated contents and is able to represent the spatial relation in them~\cite{Ruiz-2015-UPGM}. Here, an image containing complicated contents means an image that has multiple semantic information inside, such as a gastric histopathological image, which contains an admixture of the complex nuclei, cytoplasm, interstitial and tissue fluids in it. Meanwhile, annotating all abnormal (cancer) regions in gastric histopathology image is a heavy workload for the pathological 
doctors. It is necessary 
to design an efficient multi-object segmentation method. Thus, a novel Hierarchical Conditional Random field (HCRF) model is proposed 
to complete the GHIS task. Additionally, our HCRF model is developed for tissue-scale segmentation. Since the observation of histrionic optical features in 
a gastric histopathological image can be on patch-scale, and applying the entire image to 
train the classifier is not efficient and wasting computing resource, it is better to use patches of the image rather than the whole image to train the 
model~\cite{Hou-2016-PCNN}.

There are three main contributions of our work.
First of all, an HCRF framework is heuristically applied to a new research field of the GHIS. 
Secondly, a novel HCRF model using higher order potentials is proposed in this paper. 
Thirdly, high segmentation performance is obtained by the HCRF model for gastric 
histopathology images.

This paper is structured as follows: 
Sec.~\ref{s:related} summarizes related research, methodology and dataset. 
Sec.~\ref{s:CRF} illustrates how the HCRF model is built up. 
Sec.~\ref{S:exp} presents experimental results of the proposed HCRF. 
Sec.~\ref{s:discussion} compares our method to previous GHIS studies. 
Finally, Sec.~\ref{S:Con} concludes this paper with prospective work.

\section{Material and Methods}
\label{s:related}
This section is structured as follows: 
Sec.~\ref{ss:related:cancer} introduces the basic knowledge of gastric cancer.
Sec.~\ref{ss:related:image segmentation} summarizes image segmentation techniques 
in gastric histopathology field. 
Sec.~\ref{ss:related:CRF} is about the applications of CRFs. 
Sec.~\ref{ss:workflow} gives our workflow.
Sec.~\ref{ss:exp:setting} presents the dataset we use.
Sec.~\ref{ss:evaluation metrics} explains the evaluation metrics.

\subsection{Gastric Cancer}
\label{ss:related:cancer}
Gastric cancer is the amassment of abnormal cells that can be malignant, forming tumors in the stomach. It is the second most prevalent malignancy in males after 
lung cancer and the third in females after lung and breast cancer. On the basis of the 
report from WHO, nearly 800000 people die of gastric 
cancer each year~\cite{Sharma-2015-AMS}. Most of the gastric cancer cases occur in east Asia countries such as Japan and China. Particularly, the number of 
gastric cancer patients accounts for approximate $30\%$ of other types of cancer in Japan. 
Meanwhile, among countries in America, the gastric cancer cases also grow year by year~\cite{Korkmaz-2017-AES}. 
A meticulous examination of hematoxylin and eosin (H$\&$E) stained tissue slices under 
a optical microscope by pathologists is necessary during the diagnosis of gastric cancer.  
But, the microscopic examination is subjective, tedious and time-consuming. Besides, 
the screening process regularly takes 
5--10 minutes for one slide. For obtaining a high work quality, the maximum number of 
samples that a pathologist can analyse is 70 a day~\cite{Elsheikh-2013-ASO}. In case 
of leaving out any diagnostic areas, full attention are invariably required in the 
procedure. So, having a pathologist to screen and diagnose gastric cancer slides is a chief matter. With the evolution of artificial intelligence (AI), a lot of AI 
algorithms are increasingly utilized to the disease 
diagnosis~\cite{Lecun-2015-DL,Srinidhi-2019-DNN}, prompting
us to develop an intelligent diagnostic system which can effectively and accurately 
analyse gastric cancer slides~\cite{Lozano-2007-COC}.

\subsection{Image Segmentation in Gastric Histopathology Research}
\label{ss:related:image segmentation}
With the appearance of whole slide imaging technology, intelligent analysis of the whole slide histopathology image (WSI) has been accelerated. Because the manual pathological analysis by traversing the entire WSI with diverse magnifications is subjective and time-consuming owing to the large scale of WSIs (typically 100000 $\times$ 100000 pixels). The automated and accurate analysis of WSIs is promising in improving diagnostics and designing treatment strategies~\cite{Fonseca-2017-LSO}. There are three main challenges of the automatically segmenting cancerous regions in gastric WSIs: (1) The large intra-class and small inter-class variations on texture and morphology of histopathology patches make diagnosis process ambiguous. (2) Because of the considerable scale and high computing requirements of WSI, processing the entire WSI at once is difficult; and (3) the discriminative information of WSI is prone to be repressed when the abnormal regions only occupy a small proportion of the WSI compared with the normal regions~\cite{Wang-2019-RRM}.

As for the two research domains in GHIS, cell-scale 
segmentation~\cite{Sharma-2015-AMS, Wienert-2012-DSC, Kumar-2017-ADA, Cui-2018-ADL} and tissue-scale 
segmentation~\cite{Peng-2018-FCN, Li-2018-GNA,  AGC-2019-Sun,  Wang-2019-RRM, WSB-2019-Liang}, our HCRF model is developed for tissue-scale segmentation. Furthermore, an elaborate comparison of the previous GHIS studies mentioned above is shown 
in \tablename~\ref{table:discussion comparison} in Sec.~\ref{s:discussion}.

To deal with cell-scale segmentation task. In~\cite{Sharma-2015-AMS}, a multi-stage approach is proposed to segment the nuclei of 
gastric cancer from a histopathology image. 
First, a contour based minimum-model method including six main steps is used in 
the gastric cancer nuclei segmentation step~\cite{Wienert-2012-DSC}. 
Then, color, texture and morphological features are extracted from the segmented nuclei. 
Thirdly, AdaBoost, an ensemble learning method is utilized to improve the performance 
of individual classifiers. 
Lastly, the segmentation results at different resolutions are combined. Finally, an average 
multi-class classification accuracy of $58.8\%$ is obtained on twelve Her2/neu 
immunohistochemically stained gastric cancer cases which are converted to H$\&$E stain. The method in~\cite{Wienert-2012-DSC} avoids a segmentation bias and is also relatively robust against image blur. But it is time-consuming as for only segmenting nuclei.

Recently, CNN-based methods are applied to gastric cancer segmentation. 
In~\cite{Kumar-2017-ADA}, a three-class CNN algorithm is developed to segment the inside, 
outside and the boundary of gastric cancer nuclei. 
First, color normalization is applied as a pre-processing step. 
Then, a three-class CNN is designed to emphasize the nuclear boundaries. 
Lastly, three classes of segmented objects are transferred to $n$-ary nuclear maps 
for post-processing.
In the experiment, a publicly accessible H$\&$E stained dataset of histopathology images 
with more than 21000 labelled nuclear boundaries is utilized for testing. 
This dataset covers seven organs, including stomach and others. For each organ, there 
are 30 WSIs in the dataset. 
Finally, an overall F1-score of $82.67\%$ is obtained. The method in ~\cite{Kumar-2017-ADA} obtains reasonable segmentation results on organs on
which it is not trained, showing good generalization. However, the method is not robust for different magnification.

State-of-the-art networks used in GHIS are FCNs. In~\cite{Cui-2018-ADL}, an U-Net~\cite{Ronneberger-2015-UCN} based FCN is proposed to 
detect nuclei and corresponding boundaries simultaneously from gastric histopathology 
image patches. In the experiment, three datasets are tested: The first is the same 
as that used in~\cite{Kumar-2017-ADA}, and another two are breast cancer 
histopathology image datasets. Finally, F1-scores of $82.7\%$, $92.3\%$ and $84\%$ 
are achieved on these 
three datasets, respectively. The experimental results show that this method is accurate, robust and fast.

With the advancement of high speed computers, the entire abnormal tissue areas 
in the digital pathology image can be labelled automatically. In~\cite{Peng-2018-FCN}, a two-step FCN is used for gastric histopathology image classification 
and segmentation. First, patches extracted from original gastric histopathology 
images are used to fine-tune the classification CNN. Then the CNN in first step is 
transformed to a FCN for segmentation where AlexNet and GoogLeNet are used as backbones. 
Finally, segmentation accuracies of $65.7\%$ and $78.5\%$ are achieved on 400 
images, respectively. This method largely accelerates
large histopathology image analysis without damaging the accuracy too much.

To complete GHIS task, most research works focus on modifying the structure of CNNs~\cite{Li-2018-GNA, AGC-2019-Sun, Wang-2019-RRM}. In~\cite{Li-2018-GNA}, a network using multi-scale blocks which are similar to 
Inception architecture for shallow layers and feature pyramid for deep layers is 
proposed for GHIS. Outstanding result of 
$90.88\%$ F1-score is achieved on 672 gastric histopathology images. However, although this method achieves high F1-score, many normal areas are predicted to be abnormal in the segmentation result.

In another research~\cite{AGC-2019-Sun}, a network using encoder-decoder architecture which is utilized 
for fusing both low-level and high-level features, and multi-scale modules is designed. A $91.60\%$ segmentation accuracy is achieved on a dataset including 500 carefully 
annotated gastric histopathology images. This method is accurate and efficient, but it is not robust for different magnifications and staining
methods.

For getting high performance in gastric WSI segmentation result, a two-step model is proposed in~\cite{Wang-2019-RRM}. 
In the first step, a network is designed to detect discriminative patches. In the second step, 
a recalibrated multi-instance deep learning (RMDL) is used for image-level segmentation. 
This two-step method reaches $86.5\%$ classification accuracy on 608 H$\&$E stained 
and carefully labelled WSIs. This method is general and can be extended to other cancer types based on WSIs. But it is not an end-to-end model and it is time-consuming.

In order to training on a weakly annotated dataset in GHIS, making full use of already labelled areas is necessary. In~\cite{WSB-2019-Liang}, a reiterative training strategy and a particular loss 
function are utilized for partially annotated GHIS. 
A segmentation accuracy of $91.45\%$ is achieved on 1400 roughly annotated training 
images, 400 precisely annotated validation images and 100 precisely annotated test images. This method enables the network to be
trained on weakly labelled datasets. Meantime, balancing the accuracy and speeding for the GHIS task.

As for these existing DL methods applied in the GHIS, individual characteristics or network frameworks are mostly concentrated on. Strategies used to depict the integral information are not always in attention. So, to deal with this problem, CRF, an effectual and stable algorithm, is introduced.

\subsection{Applications of CRFs}
\label{ss:related:CRF}
In recent years, in order to label and parse sequential message, CRFs are utilized 
in machine vision~\cite{He-2004-MCRF}, human language comprehension or other 
biomedical signal processing areas~\cite{Lafferty-2001-CRF}. Besides, it is effective to deploy them to tag part-of-speech, accomplish shallow 
parsing task~\cite{Sha-2003-SPCR}, recognize named entity~\cite{Settles-2004-BNER}, 
find gene and peptide critical functional regions~\cite{Chang-2015-APCR}. Moreover, they are accessible choices for other machine learning (ML) algorithms, such as the 
hidden Markov models and recurrent neural network~\cite{Zheng-2015-CRFRNN}. 

Especially, more and more reserchers apply CRFs in medical image analysis fields~\cite{Wang-2006-CMFMR, Artan-2010-PCL, Park-2011-DSIA, Mary-2012-AEDC, Kosov-2018-EMC, Li-2019-CHI}.
In~\cite{Wang-2006-CMFMR}, a probabilistic discriminative algorithm is designed to merge the contextual features within functional images. The experimental results reveal that the proposed model is a robust method to detect brain activities from real functional magnetic 
resonance imaging data. In~\cite{Artan-2010-PCL}, a segmentation algorithm combining them with a cost-sensitive structure for boosting single classifier is proposed. The experiment indicates that the previous cost-sensitive 
support vector machine (SVM) results are highly promoted by adding spatial information to the CRFs. In~\cite{Park-2011-DSIA,Mary-2012-AEDC}, a CRF architecture which can effectually extract global features in colposcopy 
images of cervical cancer neoplasia is developed. It is an extraction of the domain-specific 
characteristics of diagnosis, which are in the manner of probability. As a result, 
locating the abnormal areas is achieved basing upon the relations of tissue and 
visual information.

Due to the progress of many new DL techniques, more and more studies incorporate DL 
algorithms, particularly CNNs, with CRF models to attain higher performance in image 
classification and segmentation task. For instance, in~\cite{Kosov-2018-EMC}, an 
automatic microscopic image classification and segmentation method is introduced, which is a DeepLab based strongly supervised CRF framework using CNN feature maps. At last, an overall pixel 
accuracy of $94.2\%$ and a mean average precision of $91.4\%$ are achieved. In another 
research~\cite{Li-2019-CHI}, for boosting the performance of a single classifier, a weakly supervised multilayer hidden CRF framework with 
classical ML and novel DL techniques is designed to classify cervical cancer images, 
achieving an overall accuracy of $88\%$. 

To discriminate the large intra-class and small inter-class variations on texture and morphology of a gastric histopathology image, substantial spatial information is needed. Meanwhile CRF cannot only characterize the relationship between different pixels but also different patches. So, an HCRF framework 
with CNNs techniques is newly developed for the GHIS task in this paper.

\subsection{Workflow of HCRF}
\label{ss:workflow}
\begin{figure}[htbp!]
  \centering
  \centerline{\includegraphics[width=0.8\textwidth]{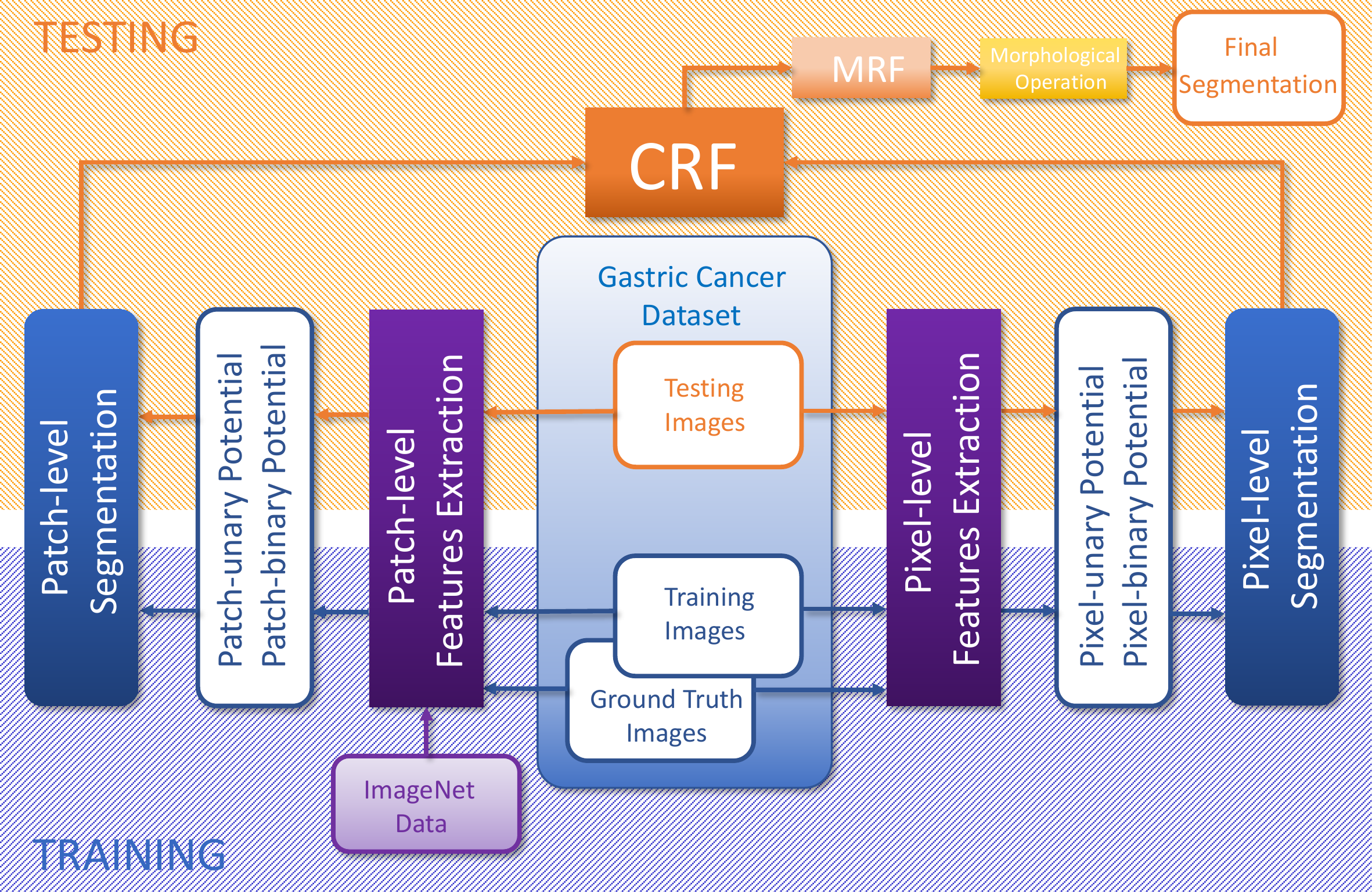}}
\caption{The workflow of the proposed hierarchical conditional random field (HCRF) model 
for gastric histopathology image segmentation (GHIS). }
\label{fig:workflow}
\end{figure}

Fig.~\ref{fig:workflow} indicates the workflow of the proposed HCRF model. The workflow is structured as below.:

First, Data Input: To train the proposed HCRF model, original 
gastric carcinoma histopathological images (original images) and corresponding 
ground truth images (GT images) are used as training and validation sets for a 
supervised learning process.

Second, HCRF Model: To build our HCRF based image segmentation 
model, four potentials are focused on, including pixel-unary, pixel-binary, 
patch-unary and patch-binary potentials.
(1) In pixel-level, pixel-level features are extracted to get the 
pixel-unary and pixel-binary segmentation results. 
(2) In patch-level, patch-level features are extracted to get three patch-unary 
and three patch-binary segmentation results. 
Then, the optimized patch-unary and patch-binary segmentation results are obtained to 
enhance the segmentation performance. (3) the main body of our HCRF image segmentation model are generated based on the pixel-unary, pixel-binary, patch-unary and patch-binary potentials.
(4) Additionally, in the segmentation result of the main body of HCRF, regions are often homogeneous and 
neighboring pixels usually have similar
properties. Meanwhile Markov Random Field (MRF)~\cite{Li-1994-MRF} is a probabilistic
model that captures such contextual
constraints. Therefore, MRF and 
morphological operations~\cite{Gonzalez-2004-DIP} are applied to optimize the 
segmentation result further.

Third, System Evaluation: To evaluate the effectiveness of the proposed HCRF 
based image segmentation method, test images are input and S\o rensen-Dice coefficient (Dice or F1-score), 
relative volume difference (RVD), 
intersection over union (IoU or Jaccard), 
precision (or positive predictive value), 
recall (or sensitivity), 
specificity (or true negative rate) 
and accuracy are calculated to measure the segmentation result, where these criteria are standard and 
suitable for medical image segmentation 
evaluation~\cite{Garcia-2017-ARO, Chang-2009-PMC, Taha-2015-MFE, Christ-2017-ALA}. 

\subsection{Experimental Settings}
\label{ss:exp:setting}
\subsubsection{Dataset}
\label{sss:exp:setting:data}
\begin{figure*}[htbp!]
  \centering
  \centerline{\includegraphics[width=0.65\linewidth]{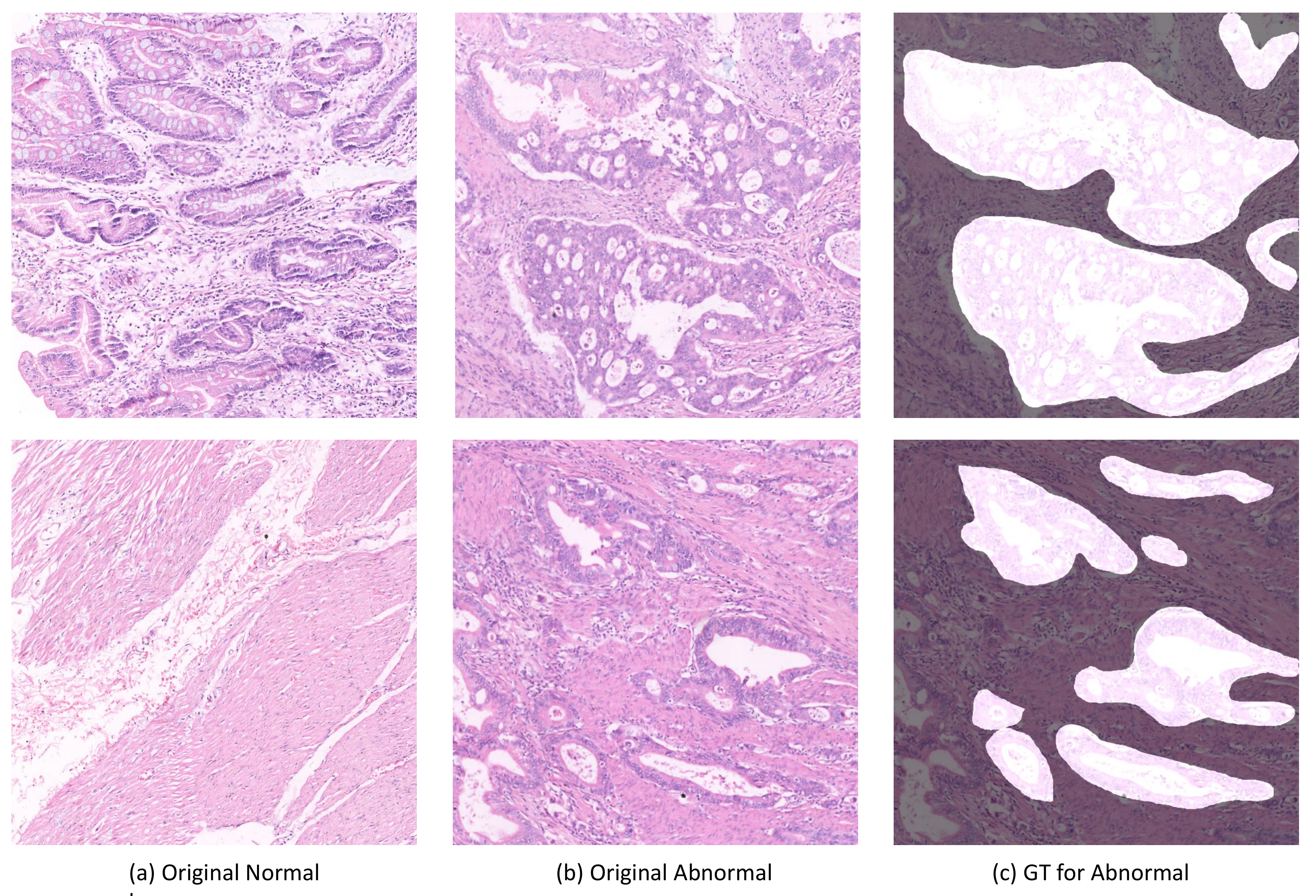}}
\caption{Examples in the H$\&$E stained gastric histopathological image dataset. 
The column (a) presents the original images of normal tissues. 
The original images in column (b) contain abnormal regions, 
and column (c) shows the corresponding GT images of the abnormal regions. 
In the GT images, the brighter regions are positive (abnormal tissues with cancer cells), 
and the darker regions are negative (normal tissues without cancer cells).}
\label{fig:db_he}
\end{figure*}

In this study, a publicly accessible H$\&$E stained gastric histopathological image dataset in 20$\times$ magnification~\cite{Li-2018-DLB} is utilized, and it is available in~\cite{Sun-2016-DFH} and represented in Fig.~\ref{fig:db_he}. In this dataset, the image format is ``*.tiff'' 
or ``*.png'' and  practical histopathologists mark most of the abnormal regions 
in histopathology images of gastric cancer. The dataset consists of 140 normal 
images without GT images, and 560 abnormal images with 560 GT images where the 
positive regions (cancerous cells) are labelled. There exists no positive regions 
in the normal images and positive regions appear in the abnormal images. The size 
of the gastric histopathological image is $2048 \times 2048$ pixels. 

\subsubsection{Training, Validation and Test Data Setting}
\label{sss:exp:setting:datasetting}
In the GHIS task, we only use the 560 abnormal images and corresponding GT 
images from the gastric histopathological image dataset. The abnormal images in the dataset
are partitioned into training, validation and test sets randomly at ratio 1:1:2. 
The dataset setting is represented in \tablename~\ref{table:data_he}. 
\begin{table}[!htbp]
\centering
\caption{Distribution of data for training, validation and testing.}
\scalebox{0.75}{
\begin{tabular}{ccccc}
\toprule
\textbf{Imge tpye} & \textbf{Training} & \textbf{Validation} & \textbf{Test} & \textbf{Sum} \\
\midrule
Original images  &  140  &  140  &  280  &  560  \\
\bottomrule
\end{tabular}}
\label{table:data_he}
\end{table}

In the pixel-level potentials, the original and GT images are flipped horizontally and 
vertically and rotated 90, 180, 270 degrees to augment the training and 
validation datasets six times. Furthermore, because the size of the gastric histopathology 
images is too large to process, the original and the GT images are cropped into 
$256 \times 256$ pixels. The information of data augmentation for the pixel-level training is shown 
in \tablename~\ref{table:data_pixel_level}.
\begin{table}[!htbp]
\centering
\caption{Data augmentation for training and validation sets in pixel-level training. 
Image types are presented in the first column. 
The usages of data are shown in the second and third columns. 
The total number of images used in this paper is presented in last column.}
\scalebox{0.8}{
\begin{tabular}{cccc}
\toprule
\textbf{Image type} & \textbf{Training} & \textbf{Validation}  & \textbf{Sum} \\
\midrule
Original images  &  140  &  140    &  280  \\

Augmented images  &  53760  &  53760  &   107520  \\
\bottomrule
\end{tabular}}
\label{table:data_pixel_level}
\end{table}

In the patch-level potentials, the original and GT images are meshed into patches 
($64 \times 64$ pixels) and the area of a patch is 4096 pixels. When the sum of the 
pixel numbers in a GT image is over 2048, a positive label (1, foreground) is given 
to the corresponding original image patch; otherwise, a negative label 
(0, background) is given to it. However, when this operation is done, the positive image patches are much 
less than the negative image patches. So, in order to balance the data during training, the positive image patches are augmented by flipping horizontally and vertically and 
rotating to 90, 180, 270 degrees in the training set. Meantime, we do the same 
augmentation of the images in the validation set. The data augmentation for the 
patch-level is shown in \tablename~\ref{table:data_patch_level}. 
\begin{table}[!htbp]
\centering
\caption{Data augmentation for training and validation sets in patch-level training. 
Image types are presented in the first column. 
The different usages of data are shown in the second and third columns. 
The bottom row presents the total number of images used in this paper.}
\scalebox{0.75}{
\begin{tabular}{ccc}
\toprule
\textbf{Image type} & \textbf{Train} & \textbf{Validation}   \\
\midrule
\multirow{1}*{Augmented positive images}  &  121251  &  119151  \\

\multirow{1}*{Augmented negative images}  &  121251  &  119151  \\

\multirow{1}*{Sum}  &  242502  &  238302     \\
\bottomrule
\end{tabular}}
\label{table:data_patch_level}
\end{table} 

\subsection{Evaluation Metrics}
\label{ss:evaluation metrics}
To give a quantitative evaluation, Dice, RVD, IoU, 
precision, 
recall, specificity and accuracy are used to measure the segmentation result. These seven criteria are defined in \tablename~\ref{table:evaluation index}.

\begin{table}[!htbp]
\renewcommand\arraystretch{2}
\centering
\caption{The seven evaluation criteria and corresponding definitions.}
\scalebox{0.75}{
\begin{tabular}{cccc}
\toprule
\textbf{Criterion} & \textbf{Definition} & \textbf{Criterion} & \textbf{Definition} \\
\midrule
\multirow{1}*{Dice}  &  $\dfrac{2\mathrm{TP}}{2\mathrm{TP}+\mathrm{FP}+\mathrm{FN}} $&  {Recall} & $ \dfrac{\mathrm{TP}}{\mathrm{TP}+\mathrm{FN}}$ \\
\multirow{1}*{RVD}  &  $ |\dfrac{\mathrm{FP}+\mathrm{TP}}{\mathrm{TP}+\mathrm{FN}}|-1$  &  {Specificity}  &  $ \dfrac{\mathrm{TN}}{\mathrm{TN}+\mathrm{FP}}$
\\
\multirow{1}*{IoU}  &  $\dfrac{\mathrm{TP}}{\mathrm{TP}+\mathrm{FN}+\mathrm{FP}}$ &  {Accuracy} & $\dfrac{\mathrm{TP}+\mathrm{TN}}{\mathrm{TP}+\mathrm{FN}+\mathrm{TN}+\mathrm{FP}}$  
\\
\multirow{1}*{Precision}  &  $ \dfrac{\mathrm{TP}}{\mathrm{TP}+\mathrm{FP}}$ &   &  
\\
\bottomrule
\end{tabular}}
\label{table:evaluation index}
\end{table}
In the definition of these criteria, TP denotes the true positive, which are positive 
cases diagnosed as positive.
TN denotes the true negative, which are negative cases diagnosed as negative. 
FP denotes the false positive, which are negative cases diagnosed as positive and FN denotes the false negative, which are positive cases diagnosed as unfavorable. 
Dice is in the interval [0,1], and a perfect segmentation yields a Dice of 1. 
RVD is an asymmetric metric, and a lower RVD means a better segmentation 
result~\cite{Christ-2017-ALA}. 
IoU is a standard metric for segmentation purposes that computes a ratio between the 
intersection and the union of two sets, and a high IoU means a better segmentation 
result~\cite{Garcia-2017-ARO}.  
Precision is utilized to measure the proportion of the relevant cases among the 
retrieved cases, where a higher precision implies a method obtains more relevant cases 
than irrelevant results substantially.
A recall is used to estimate the proportion of relevant cases that have been retrieved 
in the total number of relevant cases, where if a method obtains most of the relevant cases, 
it gets a high recall. Specificity is a metric of the ratio of actual negative cases that 
are correctly distinguished~\cite{Powers-2011-Evaluation}. 
Segmentation accuracy is ratio of the correctly predicted pixels among the total pixels, where a higher 
accuracy means a better segmentation result.

\section{Theory}
\label{s:CRF}
Firstly, Sec.~\ref{ss:CRF:basic} introduces the fundamental definition of CRFs. 
Afterwards, Sec.~\ref{ss:CRF:our} elaborates on our proposed HCRF model, 
including pixel-unary, pixel-binary, patch-unary, patch-binary potentials, 
and their combination.

\subsection{Fundamental Definition of CRFs}
\label{ss:CRF:basic}

The basic theorem of CRF is presented in~\cite{Lafferty-2001-CRF}: 
First of all, the observation sequence $\textbf{Y}$ is a random 
variable to be labelled, and $\textbf{X}$ is the random variable 
of the relative label sequence. 
Secondly, $\textit{G} = (\textit{V},\textit{E})$ is a graph where 
$\textbf{X}=(\textbf{X}_{v}) _{v\in\textit{V}}$, whilst $\textbf{X}$ is indexed 
by the nodes or vertices of $\textit{G}$. $V$ is the array of all sites, 
which corresponds with the vertices in the related undirected graph $G$, where edges 
$E$ construct the interactions among adjacent sites. Therefore, 
$(\textbf{X}, \textbf{Y})$ is a CRF 
in case, when conditioned on observation sequence $\textbf{Y}$, the random 
variables $\textbf{X}_{v}$ follow
the Markov properties with regard to the graph: 
$\textit{p}=(\textbf{X}_{v} | \textbf{Y}, \textbf{X}_{w}, w \neq v) = \textit{p}(\textbf{X}_{v} | \textbf{Y}, \textbf{X}_{w}, w \sim v)$, 
in which $w \sim v$ implies $w$ and $v$ are neighbours in $\textit{G}= (\textit{V},\textit{E})$. 
These principles indicate the CRF model is an undirected graph in which two disjoint sets 
$\textbf{X}$ and $\textbf{Y}$ are separated from the nodes. In that case, the conditional 
distribution  model is $\textit{p}(\textbf{X} | \textbf{Y})$.

Based on the definition of the random fields in~\cite{Clifford-1990-MRF}, 
the joint distribution over the label sequence $\textbf{X}$ given $\textbf{Y}$ forms 
as Eq.~\eqref{equ:1}. 
\begin{equation}
\begin{aligned}
&p_{\theta}(\textbf{x}|\textbf{y})\propto
&\textrm{exp}
(\sum_{e\in E,k}\lambda_{k}f_{k}(e,\textbf{x}|_{e},\textbf{y} )+ \sum_{v\in V,k}\mu_{k}g_{k}(v,\textbf{x}|_{v},\textbf{y} )),
\label{equ:1}
\end{aligned}
\end{equation}
where $\textbf{y}$ is the observation sequence, $\textbf{x}$ is the corresponding 
label sequence, and $\textbf{x}|_{\textit{S}}$ is the set of sections of $\textbf{x}$ 
in association with the vertices of sub-graph $\textit{S}$. 
Furthermore, from~\cite{Chen-2018-DSI, Zheng-2015-CRFRNN, Gupta-2006-CRF}, it can 
be comprehended that a redefinition of Eq.~\eqref{equ:1} is Eq.~\eqref{equ:2}.
\begin{equation}
p(\textbf{X}|\textbf{Y})=\dfrac{1}{Z}\prod_{C}\psi_{C}(\textbf{X}_{C}, \textbf{Y}),
\label{equ:2}
\end{equation}
where $Z=\sum_{\textbf{X}\textbf{Y}}P(\textbf{X}|\textbf{Y})$ is the 
normalization factor and $\psi_{C}(\textbf{X}_{C}, \textbf{Y})$ is the potential 
function over the clique $\textit{C}$. The clique $\textit{C}$ is the subset of the 
vertices in the undirected graph $\textit{G}$ , where $C \subseteq V$, 
in this way, every two different vertices are adjoining.

\subsection{Hierarchical Conditional Random Fields (HCRFs)}
\label{ss:CRF:our}

\subsubsection{The Architecture of the HCRF Model}
\label{sss:CRF:our:structure}
\begin{figure*}[htb!]
  \centering
  \centerline{\includegraphics[width=0.88\textwidth]{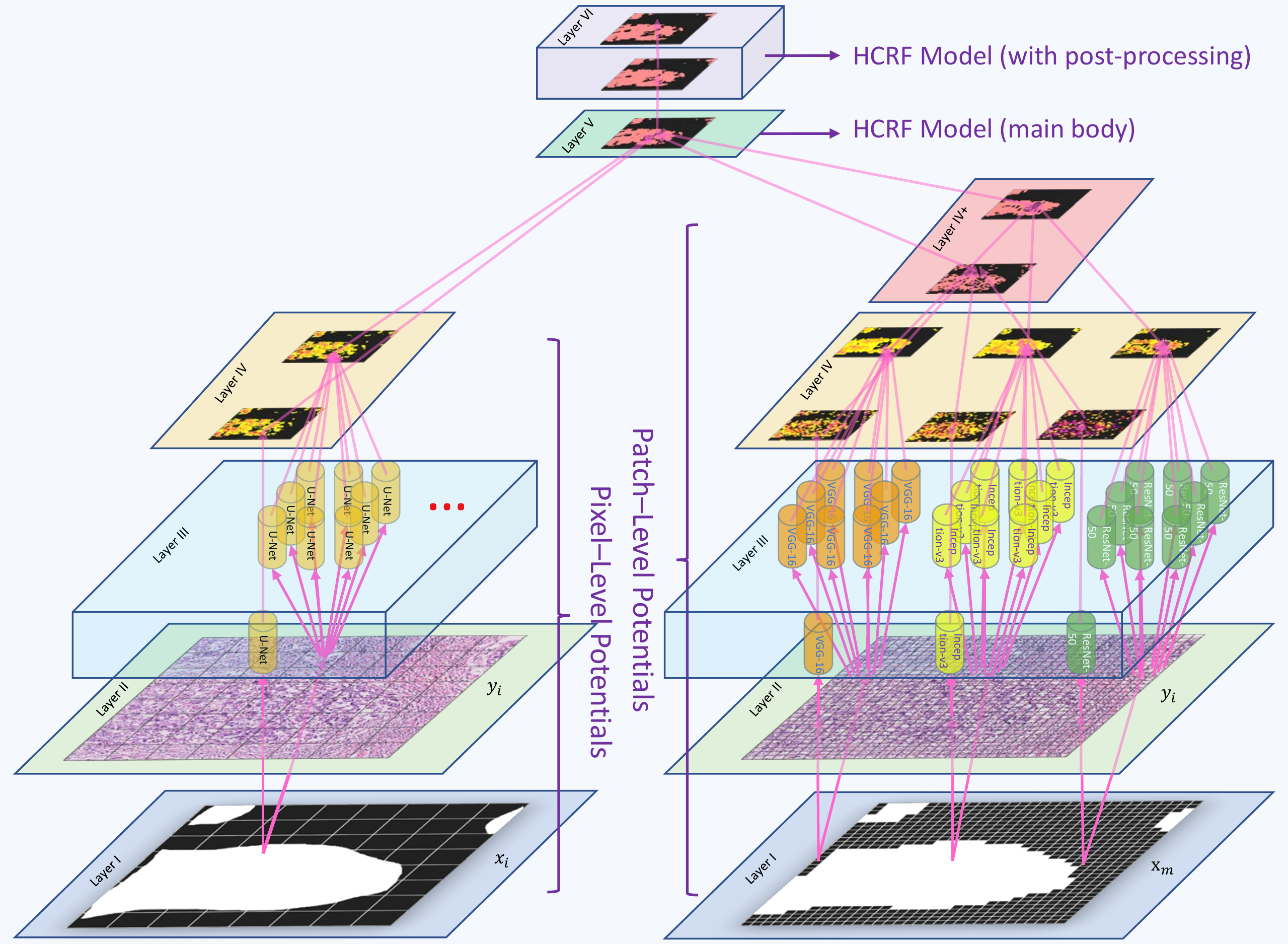}}
\caption{The architecture of the HCRF model. 
The left branch represents the terms of the pixel-level potentials, 
and the right branch represents the terms of the patch-level potentials. 
}
\label{fig:ourCRF}
\end{figure*}

Most of CRF models have been built up with only unary and binary 
potentials~\cite{Zheng-2015-CRFRNN,Chen-2018-DSI}. Nevertheless, potentials defined on 
higher order cliques have been verified to be effective in previous studies, such 
as~\cite{Vineet-2012-FBM,Arnab-2016-HOC}. Since our concentration is on the optical 
features in tissue-scale in the gastric histopathological images~\cite{Hou-2016-PCNN}, 
two types of higher order potentials are introduced. One is a patch-unary potential to 
characterize the information of tissues, the other is a patch-binary potential to depict 
the surrounding spatial relation among different tissue areas. Consequently, by the 
fundamental theorem of CRFs in Sec.~\ref{ss:CRF:basic}, our HCRF is defined as 
Eq.~\eqref{equ:3}. 
\begin{equation}
\begin{aligned}
p(\textbf{X}|\textbf{Y})=&\dfrac{1}{Z}\prod_{i \in V}\varphi_{i}(x_{i};\textbf{Y};w_{V})
\prod_{(i,j) \in E}\psi_{(i,j)}(x_{i},x_{j};\textbf{Y};w_{E})\\
&\prod_{m \in V_{P}}\varphi_{m}({\mathrm x_{m}};\textbf{Y};w_{m};w_{V_{P}})
\prod_{(m,n) \in E_{P}}\psi_{(m,n)}({\mathrm x_{m}},{\mathrm x_{n}};\textbf{Y};w_{(m,n)};w_{E_{P}}),
\label{equ:3}
\end{aligned}
\end{equation}
where 
\begin{equation}
\begin{aligned}
 Z=\sum_{\textbf{X}\textbf{Y}}\prod_{i \in V}\varphi_{i}(x_{i};\textbf{Y})\prod_{(i,j) \in E}\psi_{(i,j)}(x_{i},x_{j};\textbf{Y})
\prod_{m \in V_{P}}\varphi_{m}({\mathrm x_{m}};\textbf{Y})\prod_{(m,n) \in E_{P}}\psi_{(m,n)}({\mathrm x_{m}},{\mathrm x_{n}};\textbf{Y}),
\label{equ:4}
\end{aligned}
\end{equation}
is the normalization factor; 
$V$ represents a set of all vertices in the graph $G=(\textit{V},\textit{E})$, 
corresponding to the image pixels; 
$E$ represents a set of all edges in the graph $G$. 
$V_{P}$ is one patch divided from an image; 
$E_{P}$ represents the surrounding patches of a single patch. 
The usual clique potential function forms two components: 
The pixel-unary potential function $\varphi_{i}( x_{i},\textbf{Y})$ is the measurement of
the probability that one pixel vertex $i$ is labelled as $x_{i}\in\textbf{X}$, which gets 
values from a given set of classes $\mathbb{L}$, given input data 
$\textbf{Y}$~\cite{Kosov-2018-EMC}; 
the pixel-binary potential function $\psi_{(i,j)}(x_{i},x_{j};\textbf{X})$ is utilized 
to characterize the adjacent vertices $i$ and $j$ of the graph $G$. 
The spatial contextual relations among them are not only associated with the label of vertex 
$i$ but also to the label of its neighbouring vertices $j$. 
Furthermore, $\varphi_{m}({\mathrm x_{m}};\textbf{Y})$ and 
$\psi_{(m,n)}({\mathrm x_{m}},{\mathrm x_{n}};\textbf{Y})$ are the newly introduced 
higher order potentials. 
The patch-unary potential function $\varphi_{m}( {\mathrm x_{m}},\textbf{Y})$ is 
the measurement of the probability which a patch vertex $m$ is labelled as 
$\mathrm x_{m}$ given input data $\textbf{Y}$; 
the patch-binary potential function $\psi_{(m,n)}({\mathrm x_{m}},{\mathrm x_{n}};\textbf{Y})$ 
is utilized to present the adjacent vertices $m$ and $n$ in the patch. 
$w_{V}$, $w_{E}$, $w_{V_{P}}$ and $w_{E_{P}}$ are the weights of the four potentials, 
$\varphi_{i}( x_{i},\textbf{Y})$, 
$\psi_{(i,j)}(x_{i},x_{j};\textbf{X})$, 
$\varphi_{m}( {\mathrm x_{m}},\textbf{Y})$ and 
$\psi_{(m,n)}({\mathrm x_{m}},{\mathrm x_{n}};\textbf{Y})$, respectively. 
$w_{m}$ and $w_{(m,n)}$ are the weights of the $\varphi_{m}(\cdot;\textbf{Y})$ and 
$\psi_{(m,n)}(\cdot,\cdot;\textbf{Y})$, respectively. 
These weights are used to seek out the largest posterior label 
$\tilde{\textbf{X}}=\textrm{arg}\,\textrm{max}_{\textbf{X}}\,p(\textbf{X}|\textbf{Y})$ 
and to further improve the image segmentation performance. 
To give a visualized understanding, the architecture of the proposed HCRF model is presented 
in Fig.~\ref{fig:ourCRF}. 

From Fig.~\ref{fig:ourCRF}, the architecture of the HCRF model is as follows:
\begin{itemize}
\item Layer I presents the real labels $x_{i}$ of pixels and 
real labels $\mathrm x_{m}$ of patches in a gastric histopathology image.

\item Layer II characterizes the original image pixels $y_{i}$ that match one to one 
with the pixel labels $x_{i}$, and the original image patches $y_{i}$ that match one to 
one with the patch labels $\mathrm x_{m}$ in Layer I.

\item From the pixel-level side, Layer III represents the U-Net, which is applied in 
pixel-level potentials using images pixels $y_{i}$. To this end, the 
U-Net is trained with the gastric cancer histopathology images. From the patch-level side, Layer III denotes three types of CNNs that are applied 
in patch-level potentials 
using image patches $y_{i}$, including VGG-16, Inception-V3~\cite{Szegedy-2016-RIA} and ResNet-50~\cite{HE-2016-DRL} networks. The fine-tuning techniques, which generally indicate
retraining the pretrained CNN with the dataset corresponding
to the target task, is widely adopted in computer vision~\cite{Shin-2016-DCN, Yosinski-2014-HTR, Qu-2018-GPI}. Because training a CNN
strongly relies on initial parameters, it is significant to
obtain appropriate parameter initialization as much as
possible to prevent overfitted learning. Generally,
the early layers within a CNN are in charge of acquiring
relatively universal image features, which are considered
analogous to the conventional texture features, while the later layers are
involving more specific information corresponding to the
target task. Consequently, one may fine-tune part of or all the layers to yield
more desired results than those train from full scratch in
many cases~\cite{Shin-2016-DCN, Yosinski-2014-HTR, Qu-2018-GPI}. In the HCRF model, first, these three networks in the patch-level side are pre-trained on 1.3 million images from ImageNet 
dataset~\cite{Russakovsky-2015-ILS}. 
Then, the gastric cancer histopathology images are used to fine-tune later fully connected layers in VGG-16, 
Inception-V3 and ResNet-50 networks with a transfer learning 
strategy~\cite{Matsumoto-2016-KUN}.

\item In Layer IV, one type of pixel-unary potential is obtained, corresponding to U-Net; 
and three types of patch-unary potentials, corresponding to VGG-16, Inception-V3 
and ResNet-50 networks. 
In order to obtain the pixel-level binary potentials, the potentials 
of the surrounding pixels basing on the layout shown in Fig.~\ref{fig:48layout} are calculated. 
Similarly, to obtain the patch-level binary potentials, the potentials 
of the image patches which surround a target patch basing on the ``lattice'' 
(or ``reseau'' or ``array'') layout~\cite{Li-2019-CHI} shown in Fig.~\ref{fig:layout} are calculated. 
Particularly, there exists one additional layer in the patch-level side, namely the Layer IV+, 
where weights $w_{m}$ are given to the obtained three patch-unary potentials, respectively. 
To obtain the optimal combination of $w_{m}$, these three 
potentials are iteratively calculated to obtain the best patch-unary segmentation result. 
Similarly, weights $w_{(m,n)}$ are given to three patch-binary potentials to obtain an 
optimal patch-binary segmentation result.

\item In Layer V, first, weights $w_{V}$ and $w_{E}$ are given to two obtained pixel-level 
potentials (pixel-unary and pixel-binary potentials), respectively; and weights 
$w_{V_{P}}$ and $w_{E_{P}}$ are given to two obtained patch-level potentials (patch-unary and 
patch-binary potentials), respectively. Then, the joint probability of 
these four potentials are calculated to structure the final HCRF model.

\item In Layer VI, in order to further improve the segmentation result from Layer V, MRF and morphological operations are used as post-processing in our work. 
\end{itemize}

\subsubsection{Pixel-unary Potential}
\label{sss:CRF:our:pixel-unary}
The pixel-unary potential $\varphi_{i}(x_{i};\textbf{Y};w_{V})$ in Eq.~\eqref{equ:3} 
is related to the probability weights $w_{V}$ of a label $x_{i}$, taking a value 
$c\in\mathbb L$ given the observation data $Y$ by Eq.~\eqref{equ:5}. 
\begin{equation}
\begin{aligned}
\varphi_{i}(x_{i};\textbf{Y};w_{V})\propto{\Big(p(x_{i}=c|f_{i}(Y)\Big)^{w_{V}}}, 
\end{aligned}
\label{equ:5}
\end{equation}
where the image content is characterized by site-wise feature vector $f_{i}(Y)$ which may 
be determined by all the observation data $Y$~\cite{Kumar-2006-DRF}. The observation 
depicts a pixel whether belongs to a gastric cancer region or to the background. Especially, considering the effectiveness of U-Net in medical image segmentation tasks~\cite{Falk-2019-UND,Cciccek-2016-3dU,Ronneberger-2015-UCN}, $256 \times 256 \times 2$-dimensional pixel-level feature $F_{i}$ is used for $f_{i}(Y)$, 
obtaining feature maps at the penultimate convolution layer of the U-Net and the 
probability maps $p(x_{i}=c|f_{i}(Y)$ at the last convolution layer of the 
U-Net~\cite{Ronneberger-2015-UCN}. So, the pixel-unary potential is updated to 
Eq.~\eqref{equ:6}.
\begin{equation}
\begin{aligned}
\varphi_{i}(x_{i};\textbf{Y};w_{V})=\varphi_{i}(x_{i};\textbf{$F_{i}$};w_{V}),
\label{equ:6}
\end{aligned}
\end{equation}
where the data $Y$ determines \textbf{$F_{i}$}.

\subsubsection{Pixel-binary Potential}
\label{sss:CRF:our:pixel-binary}
The pixel-binary potential $\psi_{(i,j)}(x_{i},x_{j};\textbf{Y};w_{E})$ in 
Eq.~\eqref{equ:3} conveys how similarly the pairwise adjacent sites $i$ and $j$ is to 
take label $(x_{i},x_{j})=(c,c')$ given the data~\cite{Kumar-2006-DRF} and weights, 
and it is defined as Eq.~\eqref{equ:7}. 
\begin{equation}
\begin{aligned}
\psi_{(i,j)}(x_{i},x_{j};\textbf{Y};w_{E})\propto{}
\Big(p(x_{i}=c;x_{j}=c'| f_{i}(Y),f_{j}(Y))\Big)^{w_{E}}. 
\label{equ:7}
\end{aligned}
\end{equation}

\begin{figure}[htb!]
  \centering
  \centerline{\includegraphics[width=0.3\linewidth]{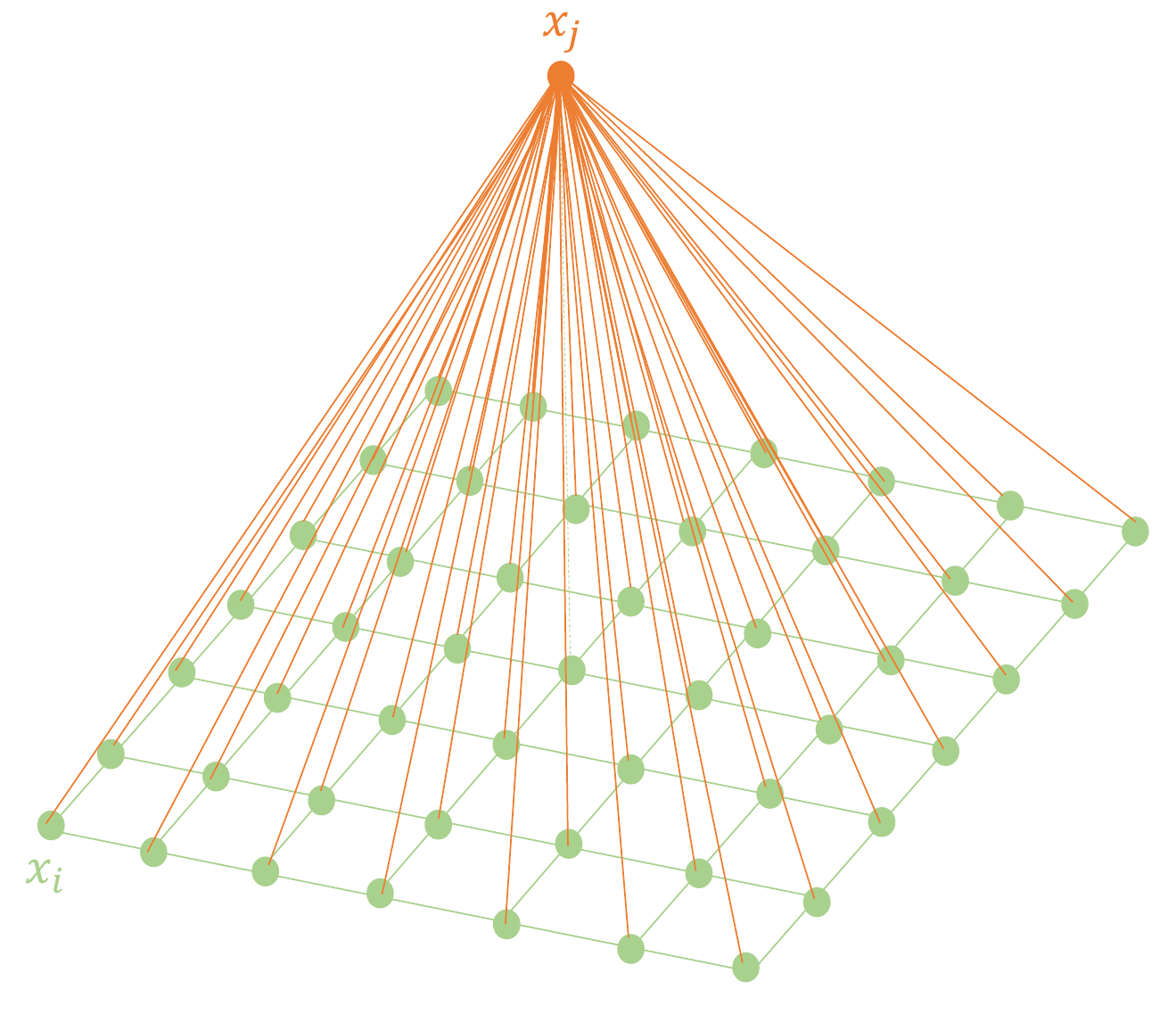}}
\caption{48 neighbourhood ``lattice'' (or ``reseau'' or ``array'') layout of the pixel-binary potential. 
The average of the unary probabilities of the 48 neighbourhood pixels is used as 
the probability of the pixel (the central pixel in orange).}
\label{fig:48layout}
\end{figure}

The layout of the pixel-binary potential is shown in Fig.~\ref{fig:48layout}. 
This ``lattice'' (or ``reseau'' or ``array'') layout is used to describe the probability of each 
classified pixel by averaging each pixel of neighbourhood unary 
probability~\cite{Li-2019-CHI}. The other procedures are identical to the pixel-unary 
potential calculation in Sec.~\ref{sss:CRF:our:pixel-unary}.

\subsubsection{Patch-unary Potential}
\label{sss:CRF:our:patch-unary}
VGG-16 uses convolution kernels with the same size to extract complex features~\cite{Simonyan-2014-VDC}. Inception-V3 is more deeper than VGG-16 
and uses multi-scale convolution kernels to extract multi-scale spatial features~\cite{Szegedy-2016-RIA}. ResNet-50 proposes deep residual learning strategy using a short-cut 
connection structure to train very deep neural 
networks~\cite{HE-2016-DRL}. In order to extract abundant spatial information, VGG-16, 
Inception-V3 and ResNet-50 networks are selected to extract patch-level features. 
In patch-level terms, $\alpha, \beta, \gamma$ are used to represent VGG-16, 
Inception-V3 and ResNet-50 networks, respectively. In patch-unary potentials 
$\varphi_{m}({\mathrm x_{m}};\textbf{Y};w_{m};w_{V_{P}})$ of Eq.~\eqref{equ:3}, 
label $\mathrm x_{m} =\{\mathrm x_{(m,\alpha)},\mathrm x_{(m,\beta)},\mathrm x_{(m,\gamma)}\}$ 
and $w_{m}=\{w_{(m,\alpha)},w_{(m,\beta)},w_{(m,\gamma)}\}$. 
$\varphi_{m}({\mathrm x_{m}};\textbf{Y};w_{m};w_{V_{P}})$ are related to the probability 
of labels $(w_{(m,\alpha)},w_{(m,\beta)},w_{(m,\gamma)})=(c,c,c)$ given the data $Y$ by 
Eq.~\eqref{equ:8}. 
\begin{equation}
\begin{aligned}
&\varphi_{m}({\mathrm x_{m}};\textbf{Y};w_{m};w_{V_{P}})\propto{}
\Big((p(\mathrm x_{(m,\alpha)}=c|f_{(m,\alpha)}(Y)))^{w_{(m,\alpha)}}\\
&(p(\mathrm x_{(m,\beta)}=c|f_{(m,\beta)}(Y)))^{w_{(m,\beta)}}
(p(\mathrm x_{(m,\gamma)}=c|f_{(m,\gamma)}(Y)))^{w_{(m,\gamma)}}\Big)^{w_{V_{P}}},
\label{equ:8}
\end{aligned}
\end{equation}
where the characteristics in image data are transformed by site-wise feature vectors 
$f_{(m,\alpha)}(Y)$, $f_{(m,\beta)}(Y)$ and $f_{(m,\gamma)}(Y)$ that may be determined by 
all the input data $Y$. For $f_{(m,\alpha)}(Y)$, $f_{(m,\beta)}(Y)$, and $f_{(m,\gamma)}(Y)$, 
we use 1024-dimensional patch-level bottleneck features $F_{(m,\alpha)}$, 
$F_{(m,\beta)}$ and $F_{(m,\gamma)}$, obtained from pre-trained VGG-16, Inception-V3 
and ResNet-50 by ImageNet; and retrain their last three fully connected 
layers~\cite{Kermany-2018-IMD} 
using gastric histopathology images to obtain the classification 
probability of each class. So, the patch-unary potential is updated to 
Eq.~\eqref{equ:9}. 
\begin{equation}
\begin{aligned}
\varphi_{m}({\mathrm x_{m}};\textbf{Y};w_{m};w_{V_{P}})=
\varphi_{m}({\mathrm x_{m}};F_{(m,\alpha)};F_{(m,\beta)};F_{(m,\gamma)};w_{m};w_{V_{P}}),
\label{equ:9}
\end{aligned}
\end{equation}
where the data $Y$ determines $F_{(m,\alpha)}$, $F_{(m,\beta)}$ and $F_{(m,\gamma)}$.

\subsubsection{Patch-binary potential}
\label{sss:CRF:our:patch-binary}
\begin{figure}[htb!]
  \centering
  \centerline{\includegraphics[width=0.23\linewidth]{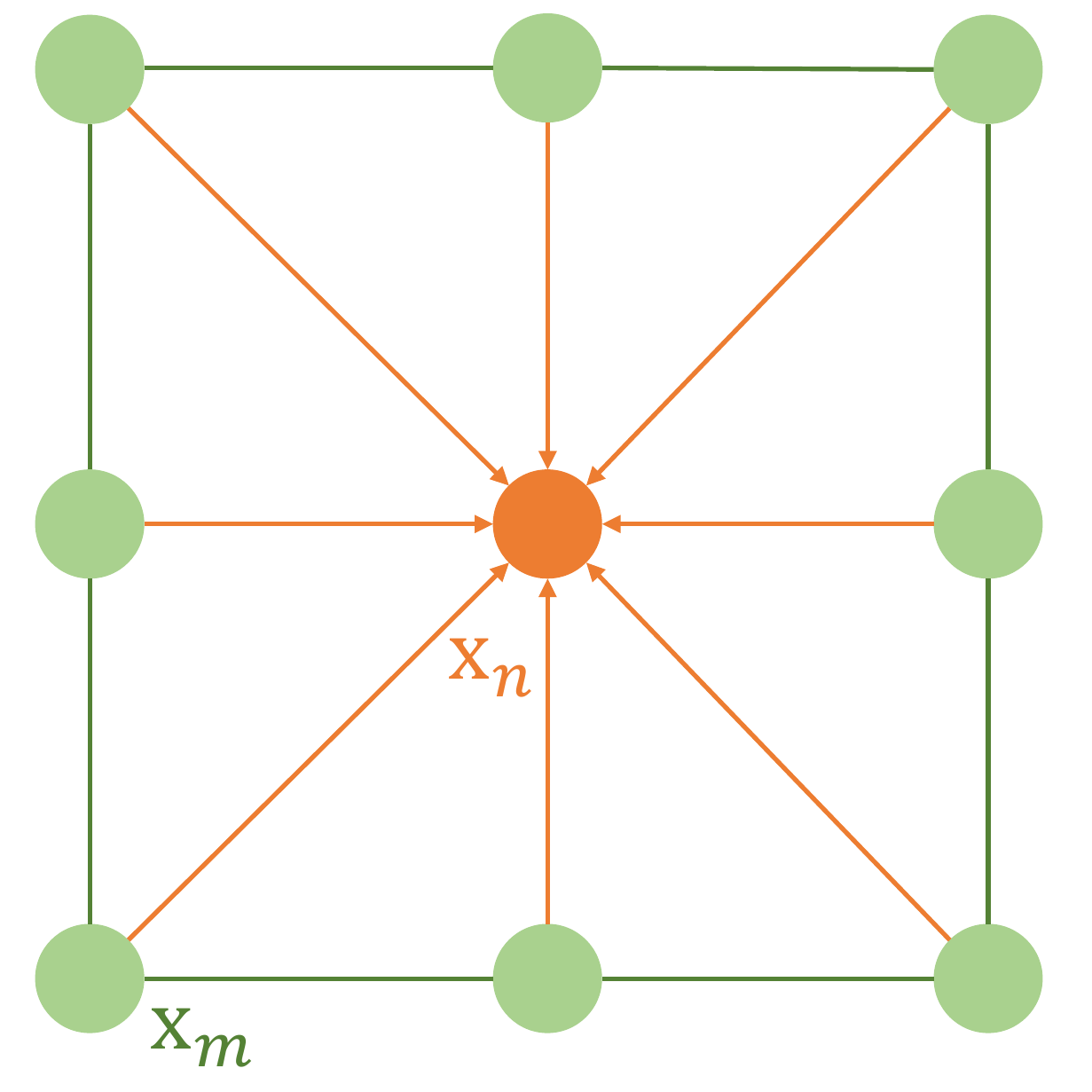}}
\caption{Eight neighbourhood ``lattice'' (or ``reseau'' or ``array'') layout of the 
patch-binary potential. 
The average of unary probabilities of the eight neighbourhood patches is utilized as 
the probability of the central target patch.}
\label{fig:layout}
\end{figure}

The patch-binary potential 
$\psi_{(m,n)}({\mathrm x_{m}},{\mathrm x_{n}};\textbf{Y};w_{(m,n)};w_{E_{P}})$ of the 
Eq.~\eqref{equ:3} denotes how similarly the pairwise adjacent patch sites $m$ and $n$ 
is to take label $(\mathrm x_{m},\mathrm x_{n})=(c,c')$ given the data and weights, 
and it is defined as Eq.~\eqref{equ:10}. 
\begin{equation}
\begin{aligned}
&\psi_{(m,n)}({\mathrm x_{m}},{\mathrm x_{n}};\textbf{Y};w_{(m,n)};w_{E_{P}})\propto{}\\
&\Big((p(\mathrm x_{(m,\alpha)}=c;\mathrm x_{(n,\alpha)}=c'|f_{(m,\alpha)}(Y),f_{(n,\alpha)}(Y)))^{w_{(m,n,\alpha)}}\\
&(p(\mathrm x_{(m,\beta)}=c;\mathrm x_{(n,\beta)}=c'|f_{(m,\beta)}(Y),f_{(n,\beta)}(Y)))^{w_{(m,n,\beta)}}\\
&(p(\mathrm x_{(m,\gamma)}=c;\mathrm x_{(n,\gamma)}=c'|f_{(m,\gamma)}(Y),
f_{(n,\gamma)}(Y)))^{w_{(m,n,\gamma)}}\Big)^{w_{E_{P}}},
\label{equ:10}
\end{aligned}
\end{equation}
where 
$\mathrm x_{n}=\{\mathrm x_{(n,\alpha)},\mathrm x_{(n,\beta)},\mathrm x_{(n,\gamma)}\}$ 
denotes the patch labels and 
$ w_{(m,n)}=\{w_{(m,n,\alpha)},w_{(m,n,\beta)},w_{(m,n,\gamma)}\}$
represents the patch weights. 
A ``lattice'' (or ``reseau'' or ``array'') layout in Fig.~\ref{fig:layout} is designed to 
calculate the probability of each classified patch by averaging each patch of neighbourhood 
unary probability~\cite{Li-2019-CHI}. The other operations are similar to the patch-binary 
potential in Sec.~\ref{sss:CRF:our:patch-unary}. 

\section{Results}
\label{S:exp}

\subsection{Evaluation of Pixel-level Potentials}
\label{ss:exp:pixel-level}

\subsubsection{Evaluation for Pixel-level Segmentation Results}
\label{sss:exp:pixel-level:result}
First, training and validation sets in 
\tablename~\ref{table:data_pixel_level} are used to train the U-Net. The validation set is 
applied to tune the CNN parameters and avoid the overfitting or underfitting of CNN during 
the training process. 
Second, the prediction probability of each pixel in an image is obtained in the validation 
and test sets. 
Thirdly, in order to evaluate the segmentation performance, the 
$256 \times 256$ pixel images are split joint to $2048 \times 2048$ pixel images, and 140 
validation images of $2048 \times 2048$ pixels and 280 test images of 
$2048 \times 2048$ pixels are used to evaluate the segmentation result.

\begin{figure}[htbp!]
  \centering
  \centerline{\includegraphics[width=0.98\linewidth]{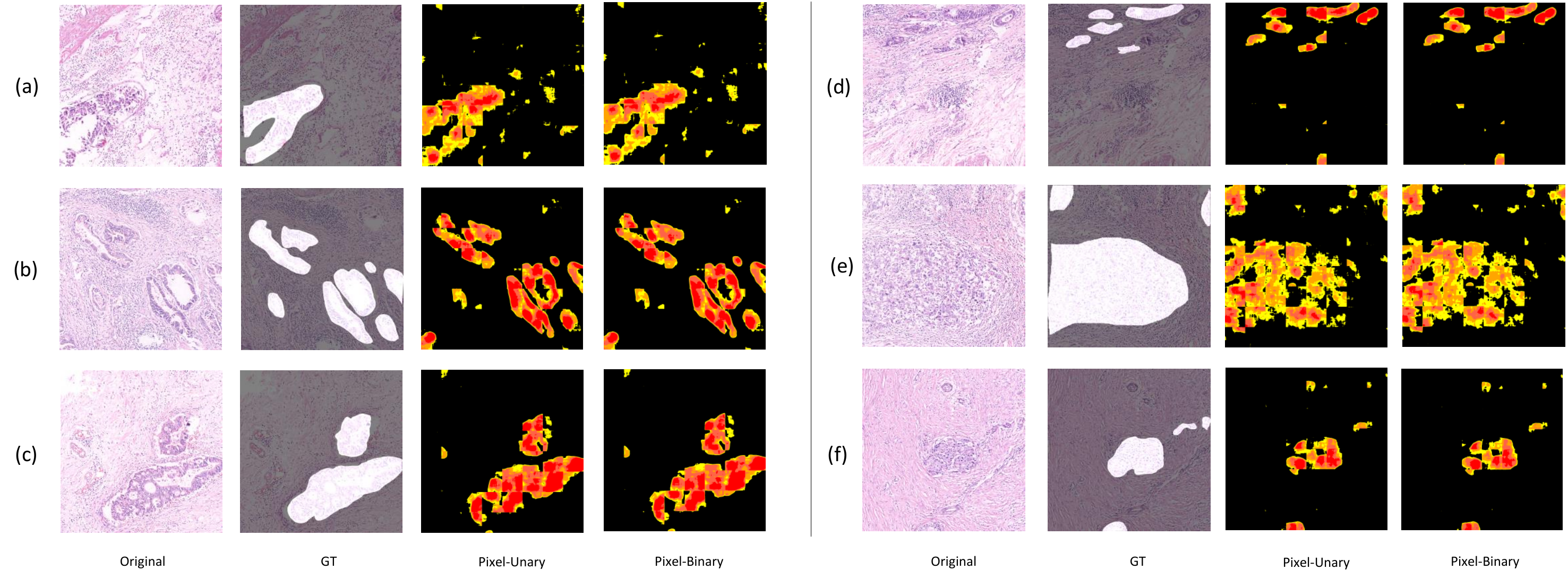}}
\caption{Examples of the pixel-level segmentation results on the validation set. 
The original and their GT images are respectively presented in the first and second columns. 
The third and last columns are the image segmentation results with pixel-unary and 
pixel-binary potentials, respectively. 
(a) is an image with a single abnormal region.
(b) and (c) are examples of multiple abnormal regions. 
(d) represents the case with very small abnormal regions. 
(e) is an example with a very big abnormal region.
(f) shows an inconspicuous case of abnormal regions. }
\label{fig:val_pixel}
\end{figure}

In Fig.~\ref{fig:val_pixel}, examples of the segmentation results 
on the validation set are shown, where different colors are used to describe different predicted 
probabilities. The 
probabilities are the higher, the regions have a greater risk of containing cancer 
tissues inside. The black, yellow, orange, tomato, red and purple colors represent 
the probabilities between $[0,0.5)$, $[0.5,0.6)$, $[0.6,0.7)$, $[0.7,0.8)$, 
$[0.8,0.9)$ and $[0.9,1]$, respectively. 

\begin{figure}[htbp!]
  \centering
  \centerline{\includegraphics[width=0.8\linewidth]{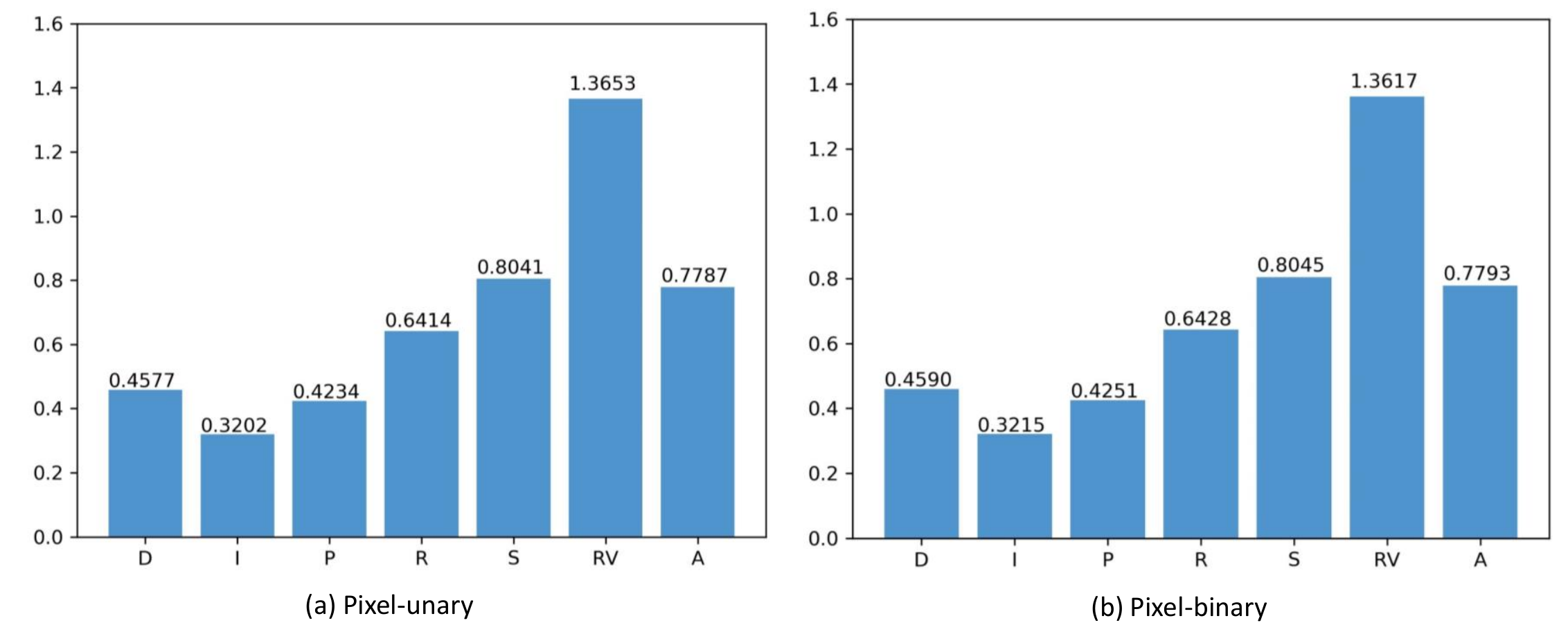}}
\caption{The evaluation for the image segmentation performance of the pixel-level 
potentials on the validation set. 
(a) and (b) are the evaluations of pixel-unary and pixel-binary potentials on the 
validation set, respectively. 
D, I, P, R, S, RV and A represent Dice, IoU, precision, recall, specificity, RVD 
and accuracy, respectively.}
\label{fig:pixel_val_evaluation}
\end{figure}

The original images in Fig.~\ref{fig:val_pixel} are common and representative cases 
in our GHIS work, so they are chosen as a visible comparison in this paper. Some noise that appears 
in the pixel-unary potential results is removed in the pixel-binary potential results, showing a strong denoising ability. However, because the binary-pixel potential makes some correctly classified pixel 
in the pixel-unary potential go wrong, the pixel-unary potential still makes sense 
in HCRF model with a complementary function. 
Meanwhile, the evaluation indexes are shown in 
Fig.~\ref{fig:pixel_val_evaluation}. 

\subsection{Evaluation of Patch-Level Potentials}
\label{ss:exp:patch-level} 

\subsubsection{Evaluation for Patch-level Segmentation Results}
\label{sss:exp:patch-level:visible}
First, the VGG-16, Inception-V3 and ResNet-50 networks are trained separately, using 
the training and validation sets in \tablename~\ref{table:data_patch_level}. 
Then, the label and predicted probability of each image patch are achieved. The 
classification accuracy of 238302 image patches in the validation set is shown in 
\tablename~\ref{table:patch_val_evaluation}.
\begin{table}[!htbp]
\centering
\caption{Classification accuracies of three CNNs on the validation set in 
patch-level training. 
The first column shows the different patch-level potentials. 
The second to the last columns show different CNNs.}
\scalebox{0.75}{
\begin{tabular}{cccc}
\toprule
\textbf{Potentials} & \textbf{VGG-16} & \textbf{Inception-V3} & \textbf{ResNet-50}  \\
\midrule
\multirow{1}*{Patch-unary}  &  0.7226  &  0.6889  &  0.7118  \\

\multirow{1}*{Patch-binary}  &  0.7331  &  0.7139  &  0.7494  \\
\bottomrule
\end{tabular}}
\label{table:patch_val_evaluation}
\end{table}

\begin{figure}[htbp!]
  \centering
  \centerline{\includegraphics[width=0.75\linewidth]{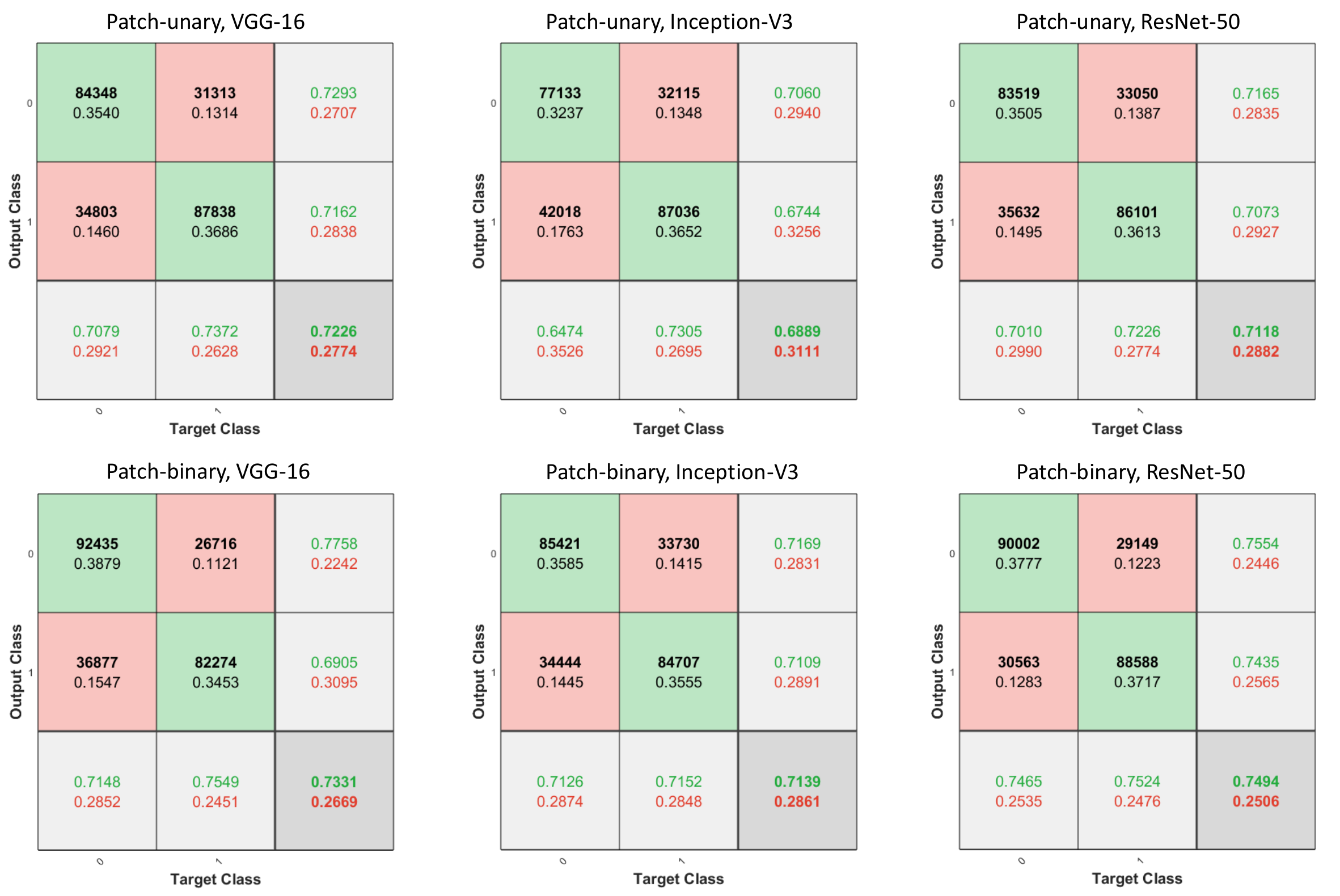}}
\caption{Image patch classification results of three CNNs on the validation set. 
The confusion matrices in the left, middle and right columns present the classification 
results of VGG-16, Inception-V3 and ResNet-50 networks of patch-unary and -binary 
potentials, respectively, where 1 represents positive and 0 represents negative.}
\label{fig:val_confusion_matrix}
\end{figure}

It can be found from \tablename~\ref{table:patch_val_evaluation} that the VGG-16 performs 
well both in the patch-unary and -binary potentials, showing a strong feature 
extraction ability of this network model in the histopathology image analysis work. The classification confusion matrices are shown in 
Fig.~\ref{fig:val_confusion_matrix}. 

\begin{figure}[htbp!]
  \centering
  \centerline{\includegraphics[width=0.88\textwidth]{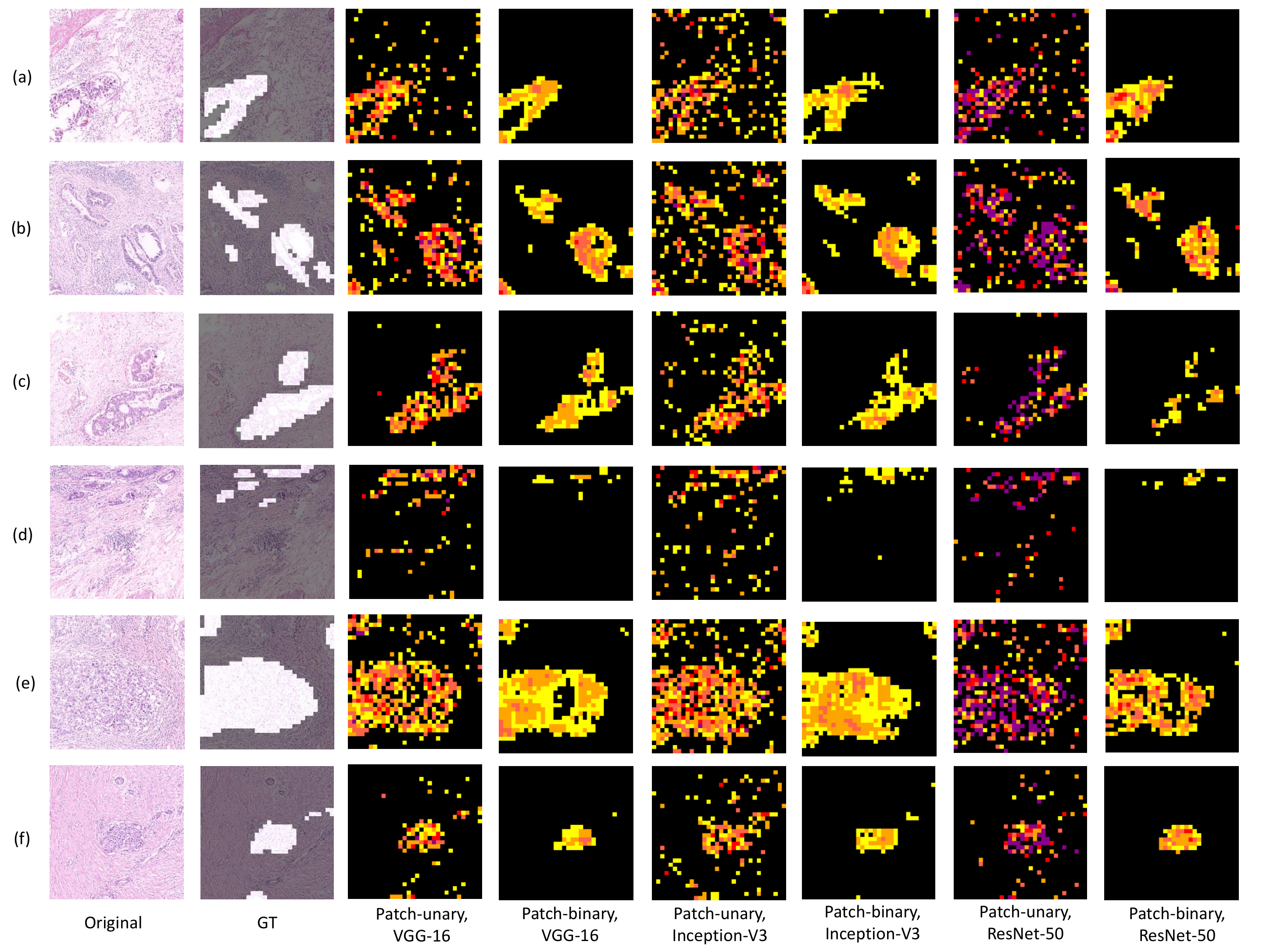}}
\caption{Examples of the patch-level segmentation results of three CNNs on the 
validation set. 
The first and second columns show the original and their GT images. 
The third and fourth columns are the VGG-16 results. The fifth and sixth columns are 
the Inception-V3 results and the seventh and last columns are the ResNet-50 results. }
\label{fig:patch_val_evaluation}
\end{figure}

Furthermore, in order to visualize the classification results, the 
patches are pieced together and the patch-level segmentation results are obtained, including three patch-unary 
results and three patch-binary results in Fig.~\ref{fig:patch_val_evaluation}. 
The probabilities of image patches corresponding to colors are set as the same as 
the pixel-level potentials in Sec.~\ref{sss:exp:pixel-level:result}.

In the patch-binary segmentation in (e) and (f) row of Fig.~\ref{fig:patch_val_evaluation}, the ResNet-50 presents better segmentation result in the main areas, but as for the upper right areas, the VGG-16 and Inception-V3 get better performance. Therefore, in order to boost performance of single CNN, these three CNNs are used simultaneously, and different weights are given to different CNNs to obtain overall better patch-level segmentation result in Sec.~\ref{sss:exp:patch-level:visible2}.  Meanwhile, from Fig.~\ref{fig:patch_val_evaluation}, it can be revealed that the patch-binary 
segmentation can eliminate the noise vary effectively. However, it may change some 
correctly segmented regions in the patch-unary potential to wrong segmentation 
results. So, both the patch-unary and patch-binary potentials make sense in our 
HCRF model.

\subsubsection{Optimization for Patch-unary and -binary Potentials}
\label{sss:exp:patch-level:visible2}
To further optimize the patch-unary and -binary segmentation results, three patch-unary potentials and three patch-binary potentials are combined
to obtain one patch-unary potential and one patch-binary potential, respectively. 
Here, two optimization strategies are compared: The first is a direct ``late fusion'' 
strategy~\cite{Snoek-2005-EVL}, where the classification probabilities of the VGG-16, 
Inception-V3 and ResNet-50 networks are given weights with a 1:1:1 ratio and 
summed together to obtain a joint classification probability. 
The second is a grid optimization strategy~\cite{Gngber-1986-GOF}, where, based on 
our pre-tests, a step length of $0.05$ is applied to give independent weights to the 
classification probabilities of the VGG-16, Inception-V3 and ResNet-50 networks. Additionally, in order to guarantee the image patch classification accuracy, the log operation is carried 
out to the probability. When the joint probability is calculated, 
this operation leads to a nonlinear case. So, in the optimized patch-level 
image segmentation results, only one pinkish-orange color is used to represent the 
positive regions and another black color to represent the negative regions.

\noindent \paragraph{\textbf{Optimization for Patch-unary Potentials}}
For the patch-unary potentials, the weights by the grid optimization are shown 
in \tablename~\ref{table:patch_val_unary_weights}. 
\begin{table}[!htbp]
\centering
\caption{The patch-unary weights of three CNNs using the grid optimization.}
\scalebox{0.75}{
\begin{tabular}{cccc}
\toprule
\textbf{CNN} & \textbf{VGG-16} & \textbf{Inception-V3} & \textbf{ResNet-50}  \\
\midrule
Weight  &  0.55  &  0.25  &  0.20  \\
\bottomrule
\end{tabular}}
\label{table:patch_val_unary_weights}
\end{table}

\begin{figure}[htbp!]
  \centering
  \centerline{\includegraphics[width=0.8\linewidth]{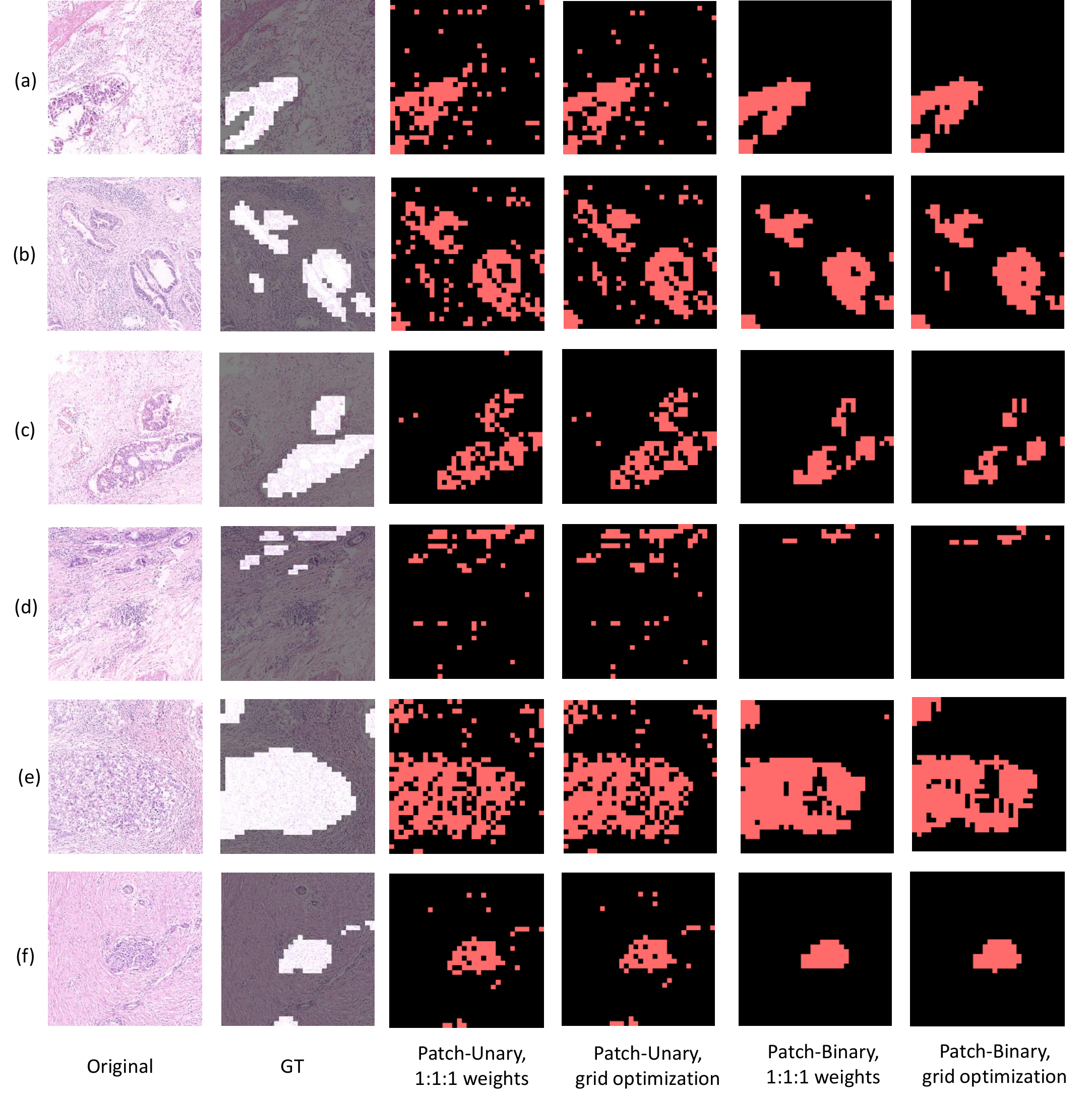}}
\caption{Examples of the optimized patch-unary and patch-binary segmentation results on 
the validation set. The first and second columns present the original and their GT images, 
respectively. 
The third and fourth columns are the optimized patch-unary image segmentation results. 
In the third column, VGG-16, Inception-V3 and ResNet-50 networks have their patch-unary 
weights with a 1:1:1 ratio. 
In the fourth column, these three CNNs have their weights as shown 
in \tablename~\ref{table:patch_val_unary_weights}. The fifth and last columns are the 
optimized patch-binary image segmentation results. 
In the fifth column, VGG-16, Inception-V3 and ResNet-50 networks have their weights 
with a 1:1:1 ratio. 
In the last column, these three CNNs have their weights as shown 
in \tablename~\ref{table:patch_val_binary_weights}.
}
\label{fig:patch_weight_evaluation-unary}
\end{figure}

It can be revealed from \tablename~\ref{table:patch_val_unary_weights} that because the VGG-16 
network has better image segmentation performance than that of other CNNs involved, 
it obtains the highest weight of 0.55. However, although the Inception-V3 and 
ResNet-50 networks have weaker performance than the VGG-16, they still contribute to 
the final segmentation results, so these two networks share the remaining weights with 0.25 and 0.20. 
Furthermore, examples of the optimized patch-unary segmentation results are shown 
in Fig.~\ref{fig:patch_weight_evaluation-unary}. 

\begin{figure}[htbp!]
  \centering
  \centerline{\includegraphics[width=0.95\linewidth]{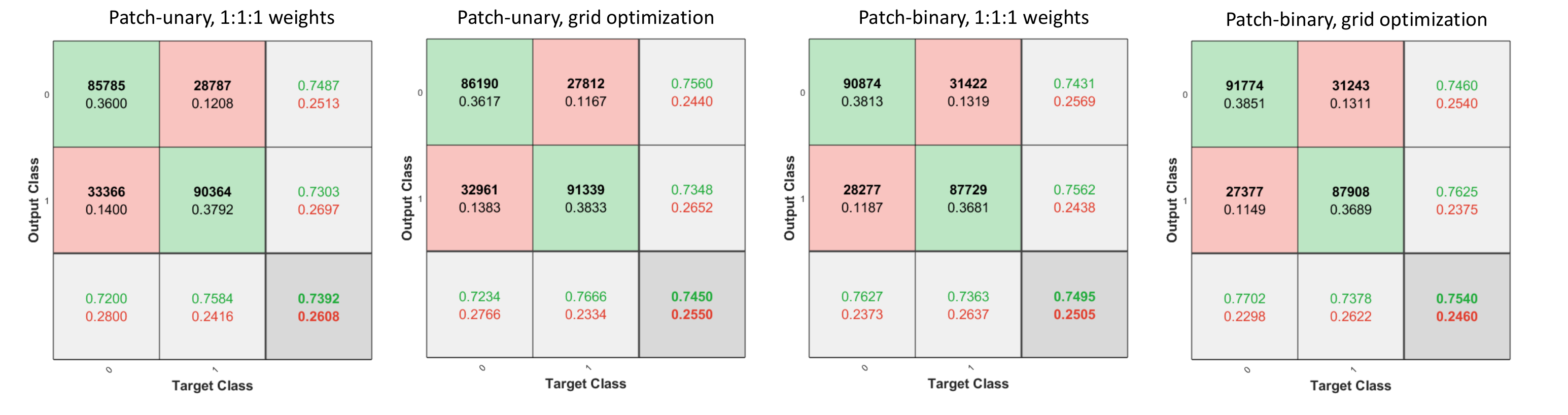}}
\caption{Image patch classification results of the optimized patch-unary and patch-binary 
potentials on the validation set. 
The first and second confusion matrices correspond to the 1:1:1 weighting and 
the grid optimization results of patch-unary potentials, respectively. The third and last 
confusion matrices correspond to the 1:1:1 weighting and 
the grid optimization results of patch-binary potentials, respectively.
1 represents positive and 0 represents negative. }
\label{fig:val_weights_confusion_matrix-patch-unary}
\end{figure}

In Fig.~\ref{fig:patch_weight_evaluation-unary}, the grid 
optimization approach achieves better image segmentation performance in details of 
the cancer regions. For more details, a comparison with the image patch 
classification confusion matrices is given in Fig.~\ref{fig:val_weights_confusion_matrix-patch-unary}. 

From Fig.~\ref{fig:val_weights_confusion_matrix-patch-unary}, we can figure out the grid 
optimization achieves a higher classification accuracy than that of the direct late 
fusion approach on both the foreground ($0.7560$) and background ($0.7348$) patches, 
as well it obtains a higher overall accuracy ($0.7450$) for all patches.

\noindent \paragraph{\textbf{Optimization for Patch-binary Potentials}} For the patch-binary potentials, 
the weights by the grid optimization is shown 
in \tablename~\ref{table:patch_val_binary_weights}. 
\begin{table}[!htbp]
\centering
\caption{The patch-binary weights of three CNNs using the grid optimization.}
\scalebox{0.75}{
\begin{tabular}{cccc}
\toprule
\textbf{CNN}& \textbf{VGG-16} & \textbf{Inception-V3} & \textbf{ResNet-50}   \\
\midrule
Weight  &  0.40  &  0.00  &  0.60  \\
\bottomrule
\end{tabular}}
\label{table:patch_val_binary_weights}
\end{table}

From \tablename~\ref{table:patch_val_binary_weights}, it can be indicated that ResNet-50 and VGG-16 networks have more robust image segmentation performance than 
that of the Inception-V3 network, so, these two CNNs obtain 0.60 and 0.40 weights, respectively. 
However, because the Inception-V3 network has much worse performance and does not 
contribute any information to the final segmentation results, it is assigned a 0 weight. Besides, examples of the optimized patch-binary segmentation results are shown 
in Fig.~\ref{fig:patch_weight_evaluation-unary}, where the grid 
optimization strategy obtains a more clean image segmentation result. More noise outside the cancer regions is removed. 
A comparison with the image patch classification system and their confusion matrices are shown 
in Fig.~\ref{fig:val_weights_confusion_matrix-patch-unary}. It reveals the 
grid optimization for patch-binary potentials obtains a higher classification accuracy 
than that of the direct late fusion approach on the foreground ($0.7460$), 
background ($0.7625$) and all ($0.7540$) image patches.

\subsection{Evaluation of the Proposed HCRF Model}
\label{ss:exp:HCRF} 

\subsubsection{Evaluation of the HCRF and Post-processing}
\label{sss:exp:HCRF:post}
\begin{figure}[htbp!]
  \centering
  \centerline{\includegraphics[width=0.9\linewidth]{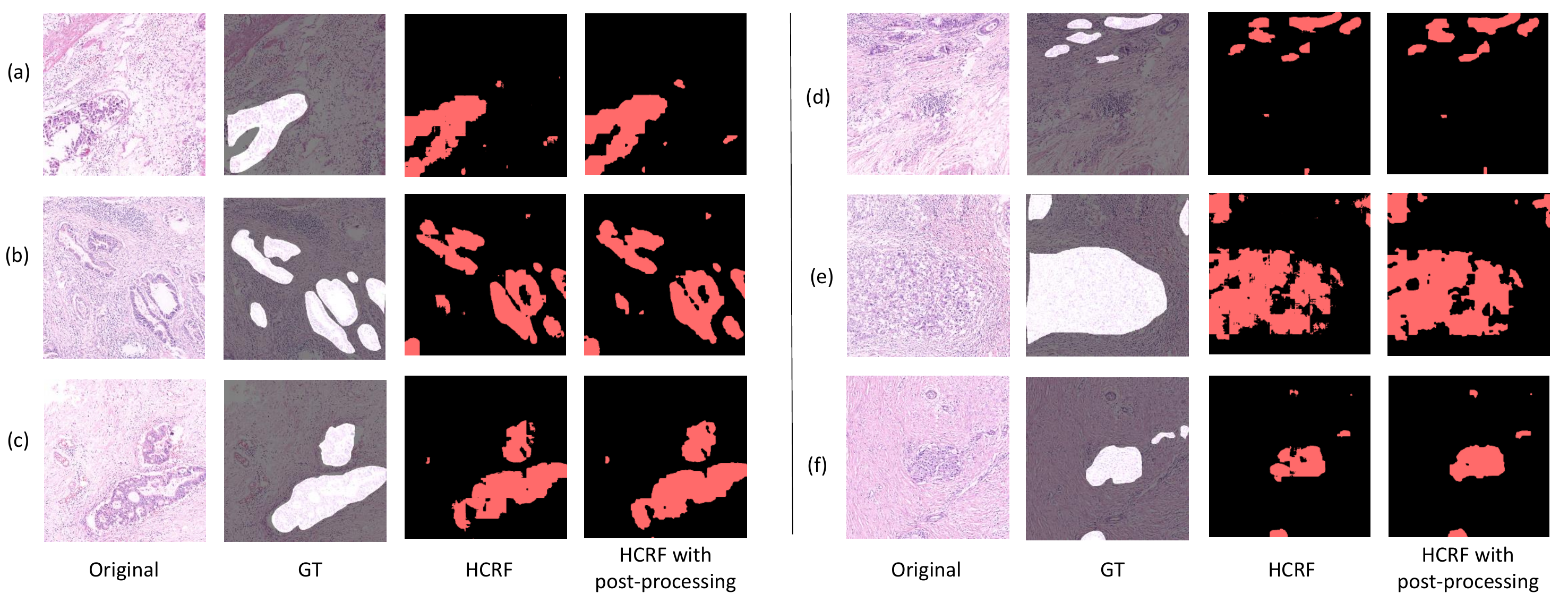}}
\caption{Examples of the HCRF image segmentation results on the validation set. 
The first and second columns present the original and their GT images, separately. 
The HCRF image segmentation results and that with the post-processing are shown in 
the third and last columns. }
\label{fig:val_final}
\end{figure}

Based on the pixel-level and patch-level image segmentation results in 
Sec.~\ref{ss:exp:pixel-level} and~\ref{ss:exp:patch-level}, the final result by the 
HCRF model is obtained, where the pixel-unary, pixel-binary, 
patch-unary and patch-binary segmentation accuracies are used as the weights to optimize 
the HCRF model. Specifically, because all of these four accuracies 
(0.7787, 0.7793, 0.7450 and 0.7540) are in $[0.7,0.8]$ and lack of discrimination, 
and considering the computation efficiency of $2048 \times 2048$ pixels, 
we use them to minus 0.7 to enhance their distinction. Hence, weights 
of 0.087, 0.0793, 0.0450 and 0.0540 are given for pixel-unary, pixel-binary, patch-unary and 
patch-binary potentials, respectively. 
In addition, the segmentation result with the post-processing 
is also achieved. Examples of the HCRF segmentation results are shown in 
Fig.~\ref{fig:val_final}. 

\begin{figure}[htbp!]
  \centering
  \centerline{\includegraphics[width=0.78\linewidth]{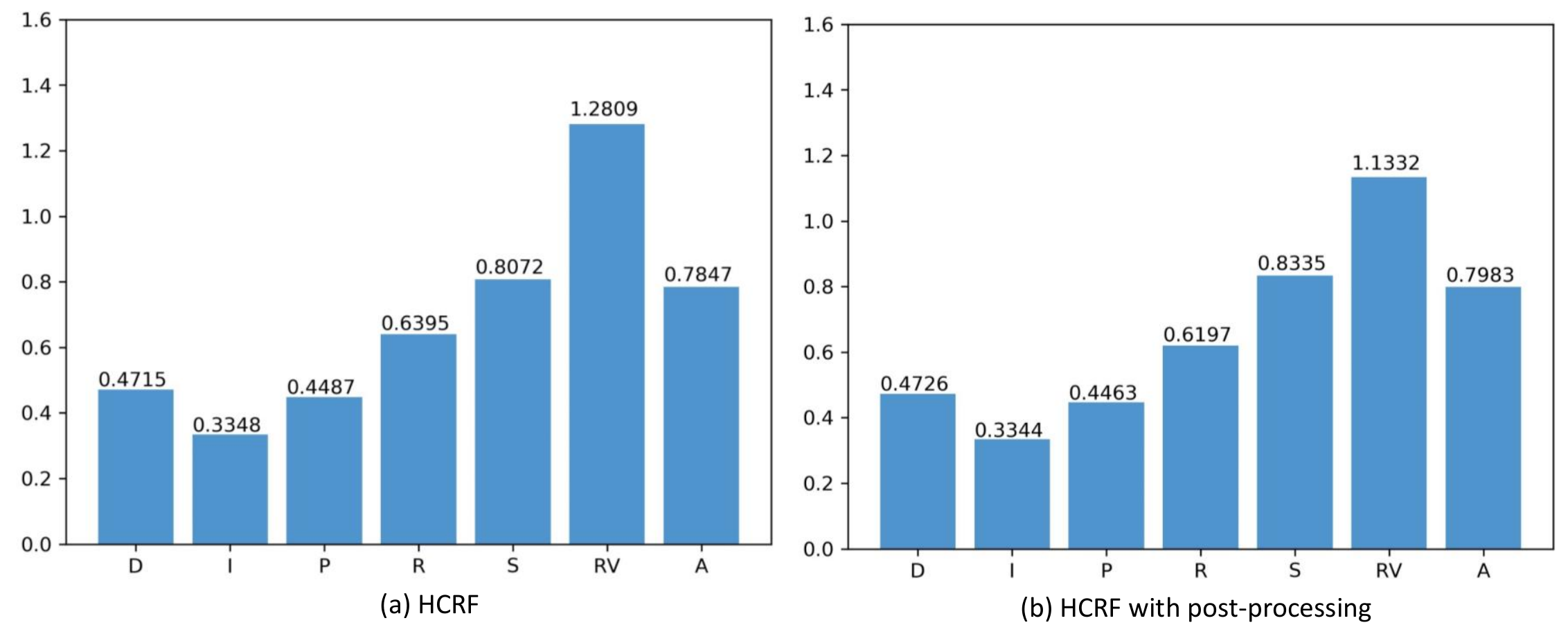}}
\caption{The evaluation for the image segmentation performance of the proposed 
HCRF model on the validation set. 
(a) and (b) are the evaluations of the proposed HCRF model and that with the 
post-processing on the validation set, respectively. 
}
\label{fig:val_final_evaluation}
\end{figure}

From Fig.~\ref{fig:val_final}, it is revealed that the HCRF model achieves better image 
segmentation performance than that of the pixel-level potentials in 
Fig.~\ref{fig:val_pixel}, where the over-segmentation and 
under-segmentation cases are reduced. Specially, when the post-processing consisted 
of two steps which are MRF and morphological operation including once open operation
is applied, the segmentation performance is further improved. In the segmentation results with post-processing of (c) and (f) row in Fig.~\ref{fig:val_final}, our post-processing connects the under-segmentation areas and smooth the edge of abnormal areas can be obviously found. In addition, the 
numerical evaluation for the HCRF segmentation performance is shown in 
Fig.~\ref{fig:val_final_evaluation}. 

From the comparison for seven evaluation criteria of the HCRF and HCRF with 
post-processing in Fig.~\ref{fig:val_final_evaluation}, it can be found that, four of 
them are improved (Dice, specificity, RVD and accuracy) and two of them are at nearly 
the same level (IoU and precision). Hence, the HCRF with post-processing has overall improved image segmentation performance, and it is chosen in our following 
experimental steps.

\subsubsection{Evaluation of the HCRF Model on the Test Set}
\label{sss:exp:HCRF:test}
\begin{figure}[htbp!]
  \centering
  \centerline{\includegraphics[width=0.85\linewidth]{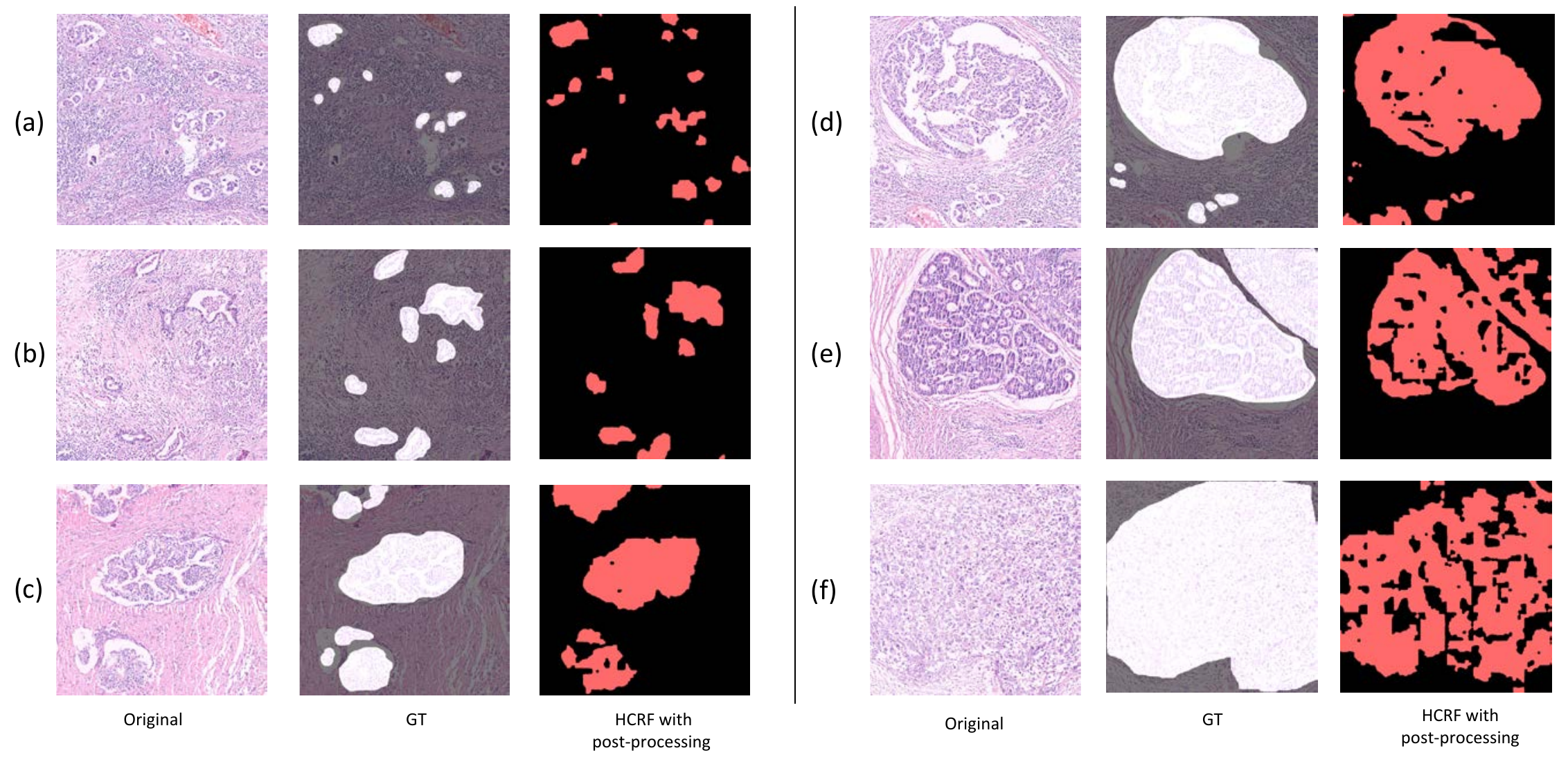}}
\caption{Examples of the HCRF segmentation results on the test set. 
The first and middle columns present the original and GT images in the test set, separately. 
The last column shows the image segmentation results by the proposed HCRF model 
(with the post-processing). 
(a) to (f) are six examples.}
\label{fig:test_evaluation1}
\end{figure}

In order to prove the effectiveness of the proposed HCRF model (with the 
post-processing), it is examined on the test set, and examples of the 
segmentation results, which include small and large targets, are presented in 
Fig.~\ref{fig:test_evaluation1}.

\begin{figure}[htbp!]
  \centering
  \centerline{\includegraphics[width=0.45\linewidth]{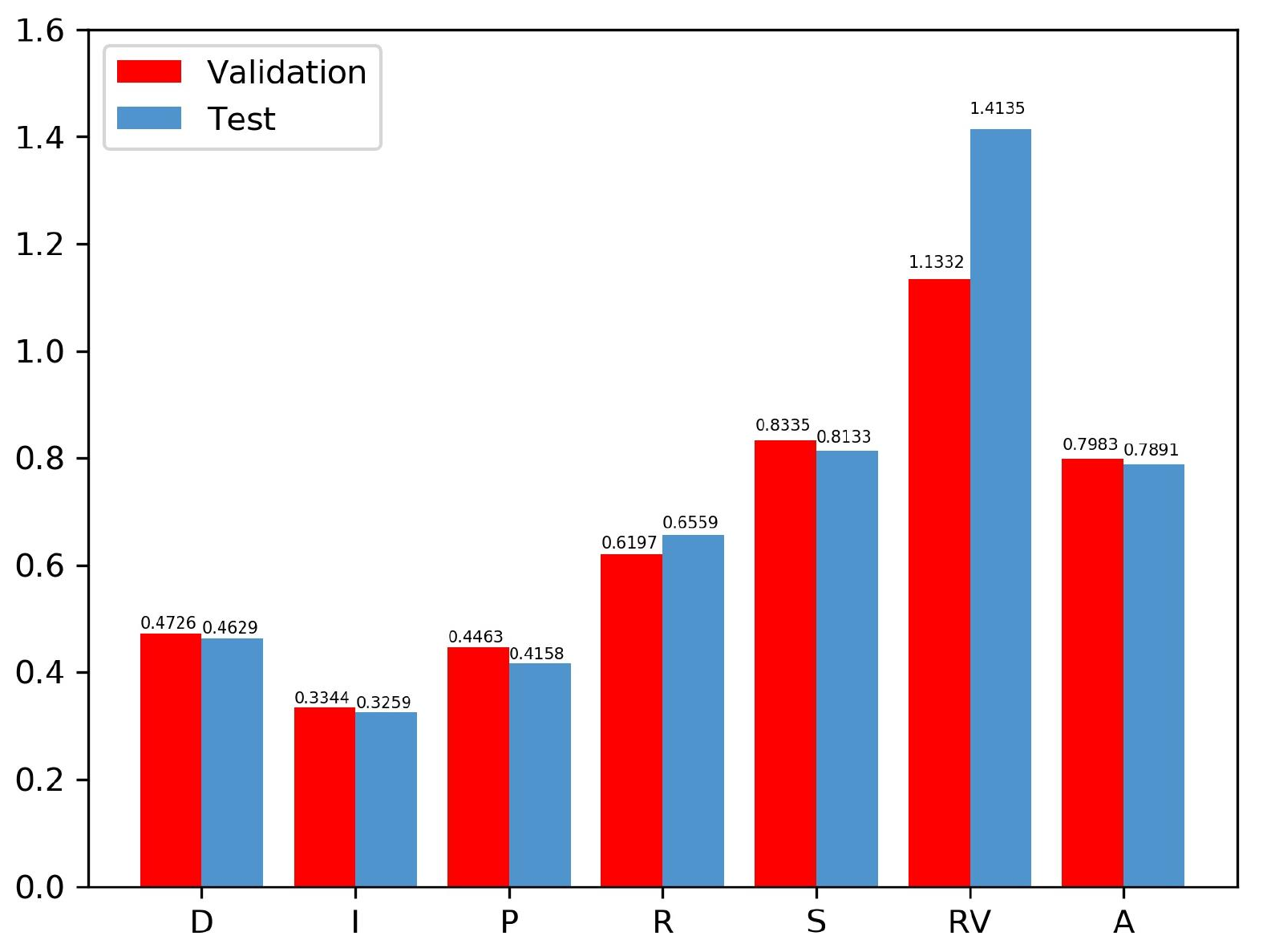}}
\caption{A comparison between the image segmentation performance of the HCRF model 
with post-processing on the validation and test sets.}
\label{fig:test_evaluation2}
\end{figure}

From Fig.~\ref{fig:test_evaluation1}, it is discovered that the HCRF model obtains good 
image segmentation results on the test set, where most of the positive (cancer) 
regions are segmented, and the edges of the regions are smooth. 
Furthermore, a numerical evaluation for the HCRF segmentation performance on 
the test set is compared with that on the validation set in 
Fig.~\ref{fig:test_evaluation2}. 

The comparison in Fig.~\ref{fig:test_evaluation2} reveals that although the 
test set has 280 images which are twice than the validation set, our proposed HCRF 
model still obtains excellent segmentation performance, where the values of all seven 
evaluation indexes on the test set are closed to that on the validation set, 
showing the high stability and a strong robustness of our method.

\subsection{Comparison to Existing Methods}
\label{ss:exp:comparison} 

\subsubsection{Existing Methods}
\label{sss:exp:comparison:methods} 
In order to show the potential of the proposed HCRF method for the GHIS task, it is compared with seven existing methods, including three state-of-the-art methods 
(DenseCRF~\cite{Chen-2018-DSI}, SegNet~\cite{Badrinarayanan-2017-SAD} and U-Net) 
and four classical methods (Otsu thresholding~\cite{Otsu-1979-ATS}, 
Watershed~\cite{Vincent-1991-WID}, $k$-means clustering~\cite{Hartigan-1979-AKC} 
and MRF). 
The experimental settings of these existing methods are briefly introduced 
as follows: (1) The DenseCRF bases on U-Net features which is trained on the dataset in 
\tablename~\ref{table:data_pixel_level} and gets $2048 \times 2048$ images finally. 
(2) The U-Net is trained on the dataset in \tablename~\ref{table:data_pixel_level} 
and gets $2048 \times 2048$ images finally. 
(3) The SegNet is trained on the dataset in \tablename~\ref{table:data_pixel_level} 
and gets $2048 \times 2048$ images finally. 
(4) The Otsu thresholding method is used on the dataset in 
\tablename~\ref{table:data_he}. 
(5) The Watershed algorithm is used with a two-stage way, where it is applied twice 
on the dataset in \tablename~\ref{table:data_he}. 
(6) The $k$-means method is trained on the dataset in 
\tablename~\ref{table:data_he}. 
(7) The MRF segmentation method is applied to the dataset in 
\tablename~\ref{table:data_he} and it includes two steps. First, we use $k$-means 
for clustering, then the MRF are applied to the $k$-means result.

\subsubsection{Image Segmentation Result Comparison}
\label{sss:exp:comparison:comparison} 
\begin{figure*}[htbp!]
  \centering
  \centerline{\includegraphics[width=0.75\textwidth]{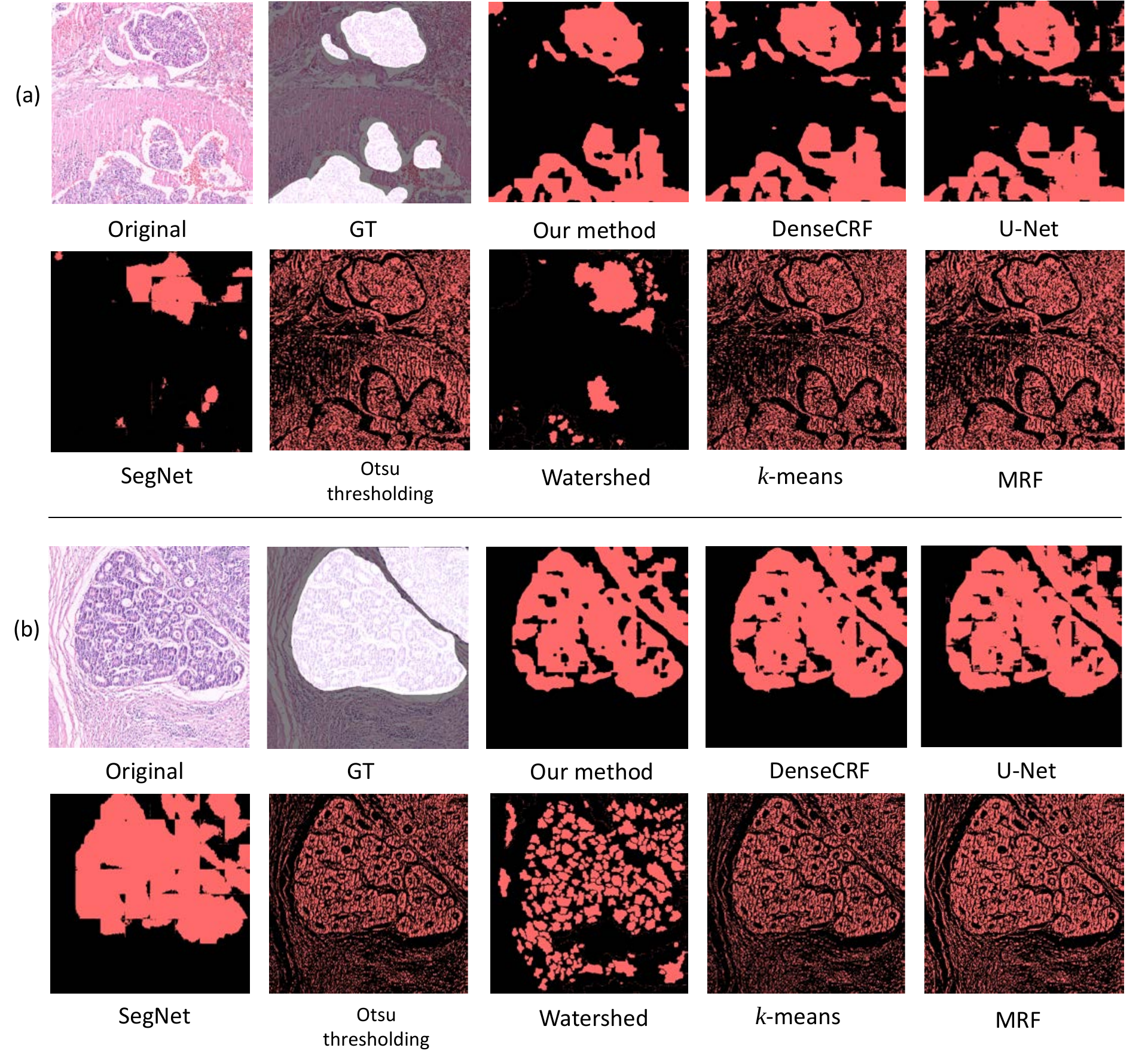}}
\caption{A comparison of the image segmentation results of our HCRF and other 
existing methods on the test set.}
\label{fig:comparison}
\end{figure*}

Fig.~\ref{fig:comparison} shows an example of the segmentation results of our method 
and seven existing methods on the test set in \tablename~\ref{table:data_he}. 

From Fig.~\ref{fig:comparison}, it can be displayed our HCRF method has better image 
segmentation performance than other existing methods in the visible comparison, 
where more positive regions (cancer) are correctly marked and less noise is remained. 
Furthermore, a numerical comparison between our HCRF method and other existing methods 
on the test set is given in \tablename~\ref{table:comparison}. 
\begin{table*}[!htbp]
\centering
\caption{A numerical comparison of the image segmentation performance between our 
HCRF model and other existing methods. 
The first row shows different methods. 
The first column shows the evaluation criteria. 
The bold texts are the best performance for each criterion.}
\scalebox{0.75}{
\begin{tabular}{ccccccccc}
\toprule
\textbf{Criterion}& \textbf{Our HCRF} & \textbf{DenseCRF} & \textbf{U-Net} & \textbf{SegNet} & \textbf{Otsu thresholding} & \textbf{Watershed} & \textbf{$k$-means} & \textbf{MRF} \\
\midrule
\multirow{1}*{Dice} & \textbf{0.4629} & 0.4578 & 0.4557 & 0.2008 & 0.2534 & 0.2613 & 0.2534 & 0.2396  \\

\multirow{1}*{IoU} & \textbf{0.3259} & 0.3212 & 0.3191 & 0.1300 & 0.1505 & 0.1585 & 0.1506 & 0.1432  \\

\multirow{1}*{Precision} & \textbf{0.4158} & 0.4047 & 0.4004 & 0.3885 & 0.2159 & 0.2930 & 0.2165 & 0.1839  \\

\multirow{1}*{Recall} & 0.6559 & 0.6889 & \textbf{0.6896} & 0.3171 & 0.4277 & 0.3541 & 0.4284 & 0.4991  \\

\multirow{1}*{Specificity} &  0.8133 & 0.7812 & 0.7795 & \textbf{0.8412} & 0.7082 & 0.7942 & 0.7078 & 0.5336  \\

\multirow{1}*{RVD} & \textbf{1.4135} & 1.6487 & 1.6736 & 2.0660 & 2.8859 & 1.9434 & 2.8953 & 4.5878  \\

\multirow{1}*{Accuracy} & \textbf{0.7891} & 0.7702 & 0.7684 & 0.7531 & 0.6598 & 0.7205 & 0.6593 & 0.5441  \\
\bottomrule
\end{tabular}}
\label{table:comparison}
\end{table*}

\tablename~\ref{table:comparison} indicates that: 
(1) Comparing to the state-of-the-art methods (DenseCRF, U-Net and SegNet), except 
recall and specificity, the proposed HCRF performs better on other indexes. The precision 
has more effectiveness in evaluating the foreground segmentation result and recall has 
more effectiveness in evaluating the background segmentation result. When focusing on 
optimizing the foreground (the positive or abnormal regions), lower FP and 
higher precision are achieved. Meantime the recall is opposite to precision. However, the Dice is 
a balance between precision and recall, and our HCRF obtains the highest Dice value, 
showing overall better image segmentation performance. 
Form Fig.~\ref{fig:comparison}, it can be indicated that the SegNet may classify some 
abnormal areas normally, leading to high TN, meantime, high specificity. 
(2) Comparing to classical methods (Otsu thresholding, Watershed, $k$-means clustering 
and MRF), our HCRF model has better segmentation results. These classical methods have 
similar results, where entire abnormal areas cannot be separated, but get more 
unnecessary details in both normal and abnormal areas.

\subsection{Computational Time}
\label{ss:exp:compute time}
At last, the computational time of our HCRF model is concisely depicted. 
A workstation equipped with Windows 10, Intel$^{ \circledR }$ Core$^{TM}$ i7-8700k 
CPU with 3.20GHz, GeForce RTX 2080 with 8GB and 32GB RAM is utilized in the experiment. The Matlab of R2018a is utilized to do the pre-processing of original image, build an architecture of HCRF, and implement segmentation result post-processing. When training the U-Net and fine-tuning the VGG-16, Inception-V3 and ResNet-50 networks, Keras framework of Python 3.6 are used as fore-end and Tensorflow framework of Python 3.6 are deployed as back-end~\cite{Chollet-2015-keras}.
\tablename~\ref{table:time} shows the training time of the U-Net and fine-tuning time of the VGG-16, Inception-V3 and ResNet-50 networks. Moreover, our dataset which has already been divided to training, validation and test set as used in this paper, and core codes of HCRF has been uploaded in ~\cite{Sun-2016-DFH}. So, our method can be reproduced when someone flows the details presented in this paper. 
\begin{table}[!htbp]
\centering
\caption{The training time of four CNNs using 140 training and 140 validation images.}
\scalebox{0.75}{
\begin{tabular}{ccccc}
\toprule
\textbf{CNN} & \textbf{U-Net} & \textbf{VGG-16} & \textbf{Inception-V3} & \textbf{ResNet-50} \\
\midrule
\multirow{1}*{Time/hours}  &  37.7  &  10.0  &  6.1  &  5.6  \\
\bottomrule
\end{tabular}}
\label{table:time}
\end{table}

Furthermore, \tablename~\ref{table:stage:time} shows the testing time of our HCRF 
model on 280 images ($2048 \times 2048$ pixels) within two working stages. 
The first stage is the ``Main Body'' of the HCRF model, which is the stage from 
Layer I to Layer V in Fig.~\ref{fig:ourCRF}. The second stage is the ``Post-processing'' 
stage, which denotes Layer VI in Fig.~\ref{fig:ourCRF}. If the histopathologists want faster results in the practical work, it is suggested to run the Main Body stage of the proposed HCRF model. Besides, for more accurate 
segmentation results, it is suggested to run the whole HCRF model with the post-processing stage. 
\begin{table}[!htbp]
\centering
\caption{The testing time of our HCRF on 280 gastric histopathology images.}
\scalebox{0.75}{
\begin{tabular}{cccc}
\toprule
\textbf{Time}& \textbf{Main Body} & \textbf{Post-Processing }& \textbf{Sum }\\
\midrule
\multirow{1}*{Total Time/hours}  &  2.2  &  2.7  &  4.9 \\

\multirow{1}*{Average Time/seconds}  &  28.1  &  33.5  &  61.6 \\
\bottomrule
\end{tabular}}
\label{table:stage:time}
\end{table}

\section{Discussion}
\label{s:discussion}

\subsection{Comparison between the Proposed HCRF Model and Previous Studies}
\label{ss:CBP}

Although our HCRF model achieves excellent segmentation performance on gastric histopathology images, our method are compared to the previous GHIS 
studies in \tablename~\ref{table:discussion comparison}. Because segmentation accuracy is regularly used and effective criterion, it is selected as the comparison metric. With the advancement of the machine vision 
algorithm applied in the pathology image segmentation and hardware 
computing power, the segmentation task in gastric histopathology image 
turns to tissue-scale from cell-scale~\cite{Srinidhi-2019-DNN}. The 
tissue-scale segmentation task is more difficult, because the tissue-scale 
content is more complex than the cell-scale content which only includes 
nuclei and cytoplasm. It can be found that cell-scale segmentation task 
usually needs fewer images~\cite{ Kumar-2017-ADA, Cui-2018-ADL}, because 
one gastric cancer slice includes massive separated cells for training. The most popular algorithm applied in the tissue-scale gastric image segmentation is 
FCN~\cite{Peng-2018-FCN, WSB-2019-Liang} and some involve the addition 
of modules to FCN~\cite{Li-2018-GNA, Wang-2019-RRM, AGC-2019-Sun}. Generally, these methods obtain high segmentation accuracies. Training the FCN 
for tissue-scale segmentation usually needs a well-labelled 
dataset~\cite{Li-2018-GNA, Wang-2019-RRM, AGC-2019-Sun}, for our dataset 
which is weakly-labelled, we do not concentrate on optimizing the FCN 
structures. Instead, the U-Net is embedded to our proposed HCRF model 
for extracting pixel-level features. Moreover, higher 
order information is extracted to characterize the gastric histopathology image. A high segmentation accuracy is achieved as well.
In \tablename~\ref{table:discussion comparison}, it can be discovered that 
the segmentation time for one gastric slice ranging from 5.1 seconds to 
244 seconds and our computational time is at the middle level. However, 
the computational power of different methods is hard to measure for 
the different image sizes and equipment. 
\begin{table*}[!htbp]
\centering
\caption{A comparison of the image segmentation methods in gastric histopathology studies}
\scalebox{0.75}{
\begin{tabular}{ccccccc}
\toprule
\textbf{Method}& \textbf{Segmentation scale} & \textbf{Training} & \textbf{Validation} & \textbf{Test} & \textbf{Accuracy} & \textbf{Time per slice/seconds}  \\
\midrule
\multirow{1}*{MM~\cite{Wienert-2012-DSC}} & Cell & - & - & 35 & - & $72.5\pm 18.2$  \\

\multirow{1}*{CNN3~\cite{Kumar-2017-ADA}} & Cell & 12 & 4 & 14 & - & 30  \\

\multirow{1}*{NBN~\cite{Cui-2018-ADL}} & Cell & 12 & 4 & 14 & - & 5.1  \\

\multirow{1}*{SWB (AlexNet)~\cite{Peng-2018-FCN}} & Tissue & 300 & - & 100 & 0.821 & 244.3   \\

\multirow{1}*{FCN (AlexNet)~\cite{Peng-2018-FCN}} &  Tissue & 300 & - & 100 & 0.657 & 12.5  \\

\multirow{1}*{FCN (GoogLeNet)~\cite{Peng-2018-FCN}} & Tissue & 300 & - & 100 & 0.785 & 14.8 \\

\multirow{1}*{GT-Net~\cite{Li-2018-GNA}} & Tissue & 560 & - & 112 & - & - \\

\multirow{1}*{RMDL~\cite{Wang-2019-RRM}} & Tissue & 408 & - & 200 & 0.865 & 93.79 \\

\multirow{1}*{RL~\cite{WSB-2019-Liang}} & Tissue & 1400 & 400 & 100 & 0.9145 & 11 \\
\multirow{1}*{DCMEN~\cite{AGC-2019-Sun}} & Tissue & 350 & - & 150 & 0.916 & - \\
\multirow{1}*{HCRF (our method)} & Tissue & 280 & 140 & 140 & 0.7891 & 61.6 \\
\bottomrule
\end{tabular}}
\label{table:discussion comparison}
\end{table*}

Although a publicly accessible gastric histopathology image dataset is used in this research, 
there exists barriers to compare our studies to previous studies. The datasets 
used in~\cite{Peng-2018-FCN, Li-2018-GNA} are identical with ours. However, 
these researchers only use a part of the dataset for training and testing in~\cite{Peng-2018-FCN}. 
Though less training data and more abnormal images in the test dataset is used to 
verify our segmentation model, similar segmentation performance is achieved with 
the SWB (AlexNet) and FCN (GoogLeNet) in~\cite{Peng-2018-FCN}. Because different abnormal images are used in the test dataset, which are randomly selected, there exists a bias in segmentation result evaluation. Since the abnormal areas in 
the gastric histopathology images of the dataset are not totally labelled, 
in~\cite{Li-2018-GNA}, the authors manually re-annotated the GT. Annotating the abnormal 
areas in pathology images is time-consuming. Therefore, taking full advantages 
of the already labelled areas provided in the dataset, where the higher order 
potentials for obtaining sufficient contextual information is used to build up HCRF 
model and automatically transfer the pixel-level GT to patch-level GT for a 
patch-level training, is the prior work. In~\cite{WSB-2019-Liang}, the researchers also use a partially 
labelled dataset and do not manually re-draw the GT either. Instead, a multi-around training strategy is chosen to train the CNN and fully use
already existing information in their dataset.

\subsection{Mis-segmentation Analysis}
\label{ss:MSA}
\begin{figure}[htbp!]
  \centering
  \centerline{\includegraphics[width=0.6\linewidth]{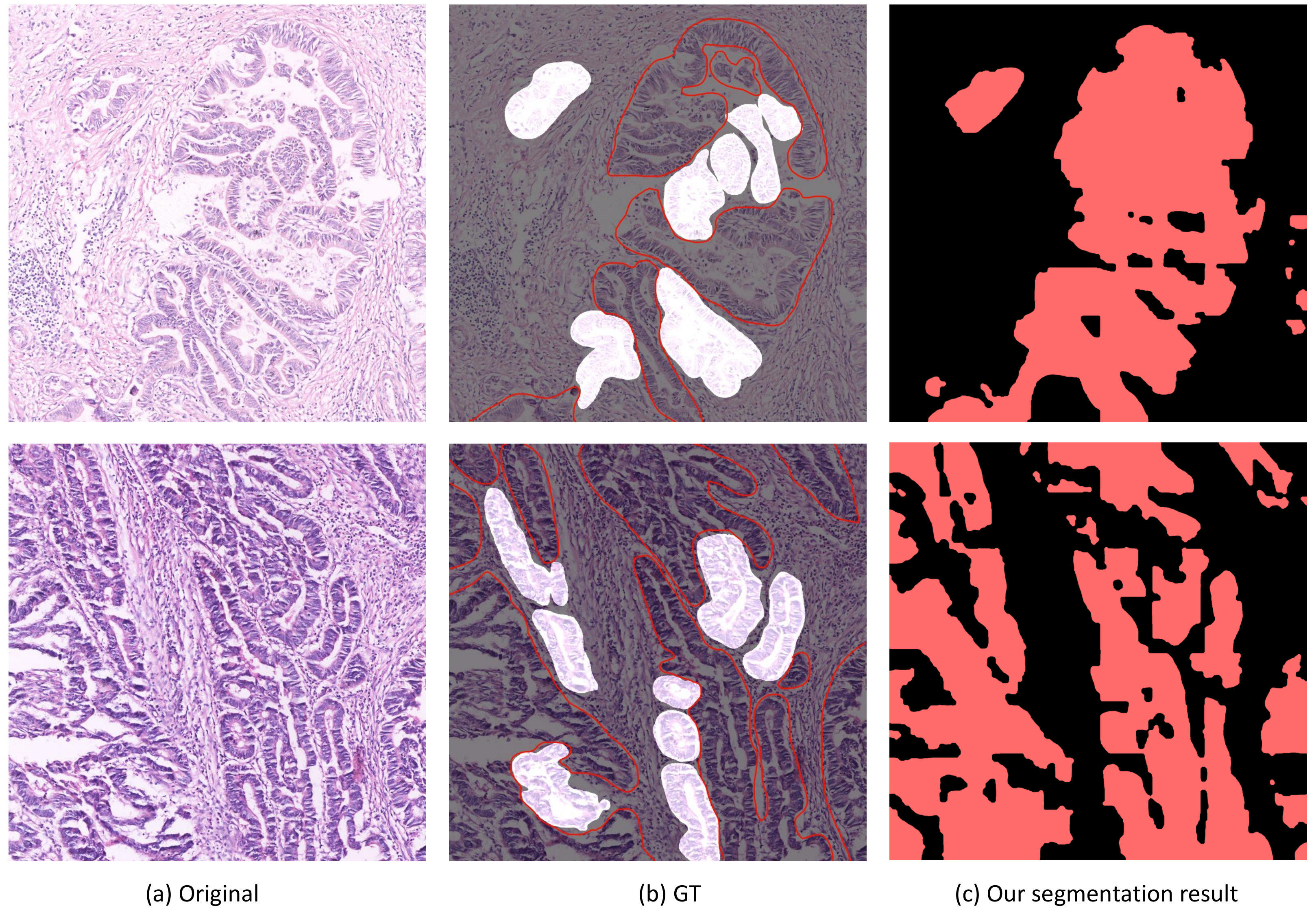}}
\caption{Typical examples of the mis-segmentation results. 
(a) presents the original images. 
(b) denotes the GT images. 
(c) shows the mis-segmentation results. 
The regions in the red curves in (b) are the positive regions in the redrawn GT images. }
\label{fig:Missegemntation}
\end{figure}

There is still a big gap of the mis-segmentation problems to overcome. 
To analyse the causes of mis-segmentation, an example is given in 
Fig.~\ref{fig:Missegemntation}. 

According to Fig.~\ref{fig:Missegemntation} and the medical knowledge from our 
cooperative histopathologists, the reasons for image segmentation errors are as 
follows: \\
(1) Fig.~\ref{fig:Missegemntation}(a) reveals that the contents of the gastric 
histopathological images are complicated, in which the characteristics between normal and 
abnormal areas are always hard to distinguish, leading to a barrier in feature 
extraction or image content description. \\
(2) As shown in Fig.~\ref{fig:Missegemntation}(b), when there are too many abnormal regions in a 
gastric histopathological image, the medical doctors draw the GT images 
roughly, where not all of the abnormal regions are figured out. This low quality 
operation makes some positive regions labelled as negative, adding a training 
difficulty.   \\
(3) From Fig.~\ref{fig:Missegemntation}(b) and (c), it can be found that, when evaluating
the segmentation results, our HCRF model may segment the positive regions correctly, 
but the original GT images may miss the corresponding regions. We consult our 
cooperative histopathologists and use red curves to redraw the regions that the 
original dataset did not label, where it can be seen that the original GT images in the 
applied dataset miss quite a lot of positive regions. It is obvious that the 
foreground of our segmentation result is closer to the redrawn GT images, 
but not the original GT images, and this case could lead to a low IoU and high SVD.

\section{Conclusions}
\label{S:Con}
In this research, an HCRF model is introduced to accomplish the GHIS task. This HCRF model not only uses traditional unary and binary potentials but also applies higher order potentials to improve the segmentation quality. 
In pixel-level potentials, the U-Net is trained; in patch-level potentials, the VGG-16, Inception-V3 and ResNet-50 networks are fine-tuned. Furthermore, when jointing the 
pixel-level and patch-level potentials, different weights for different potentials are used to optimize the model. In the experiment, our HCRF model is finally evaluated on 
a gastric H$\&$E histopathological test set and obtains Dice, IoU, precision, recall, specificity, RVD and segmentation accuracy of 46.29\%, 32.59\%, 41.58\%, 65.59\%, 81.33\%, 141.35\% and 78.91\%
which is nearly close to the 47.26\%, 33.44\%, 44.63\%, 61.97\%, 83.35\%, 113.32\% and 79.83\% on the validation set, showing 
the robustness and potential of our method.

In the future, more higher potentials will be added, such as object detection 
potentials~\cite{Arnab-2016-HOC} to improve the segmentation quality. Meanwhile, other state-of-the-art CNNs with different structures for more accurate segmentation 
results also will be accessed by our HCRF model.

\subsection*{Disclosures}
The authors declare that there are no conflicts of interest related to the research presented in this article.

\acknowledgments 
This work is supported by the ‘‘National Natural Science
Foundation of China’’ (No. 61806047), the ‘‘Fundamental
Research Funds for the Central Universities’’ (No. N2019003)
and the ‘‘China Scholarship Council’’ (No. 2017GXZ026396,
2018GBJ001757). We thank Miss Zixian Li and Mr. Guoxian Li
for their important support and discussion in this work.
\bibliography{paper}
\bibliographystyle{spiejour}

\vspace{2ex}\noindent\textbf{Changhao Sun} received his B.E. degree in communication engineering from the Northeastern University, China, in 2018. Currently, he is a Master Student in the Research Group for Microscopic Image and Medical Image Analysis in the Northeastern University, China. His research interests are gastric histopathology image segmentation, conditional random fields and deep learning. 

\vspace{1ex}\noindent\textbf{Chen Li} received his Dr.- Ing. degree from the University of Siegen (1.0 score, MAGNA CUM LAUDE), Germany in 2016. From 2016 to 2017, he worked as a Postdoctoral Researcher in the Johannes Gutenberg University Mainz, Germany. Currently, he is working as an Associate Professor in the Northeastern University, China. His research interests are microscopic image analysis, machine learning, pattern recognition, machine vision, multimedia retrieval and membrane computing. 

\vspace{1ex}\noindent\textbf{Jinghua Zhang}  received his B.E. degree from Hefei University, PR China, in 2018. Currently, he is a Master Student in the Research Group for Microscopic Image and Medical Image Analysis in the Northeastern University, China. His research interests are microscopic image segmentation and deep learning.

\vspace{1ex}\noindent\textbf{Muhammad Rahaman} received the
B.Sc. degree with BRAC University, Dhaka, Bangladesh, in 2017. He is currently pursuing the master's degree with the Research Group for
Microscopic Image and Medical Image Analysis, College of Medicine and Biological Information Engineering, Northeastern University. His
research interests are in microscopic image analysis,
medical image analysis, machine learning,
pattern recognition, and machine vision.

\vspace{1ex}\noindent\textbf{Shiliang Ai} received his B.Sc. degree from the Northeastern University, PR China, in 2016. From 2018 till now, he is a Master Student in the Research Group for Microscopic Image and Medical Image Analysis in the Northeastern University, China. His research interests are gastric histopathology image analysis and graph theory.

\vspace{1ex}\noindent\textbf{Hao Chen} received his B.E. and M.E. degrees from the Northeastern University, China, in 2017 and 2019, respectively. He is a member of the Research Group for Microscopic Image and Medical. Currently, he is pursuing for his Doctor degree. His research interests include gastric histopathology image analysis, conditional random fields, feature extraction and image classification.

\vspace{1ex}\noindent\textbf{Frank Kulwa} received his B.E. degree from the Dar es Salaam Institute of Technology, Tanzania, in 2013, where he has been a Tutorial Assistant, since 2017. Since 2018, he has been a Master Student with the Research Group for Microscopic Image and Medical Image Analysis, Northeastern University, China. His research interests include microscopic image segmentation and deep learning.

\vspace{1ex}\noindent\textbf{Yixin Li} was born in 2000. Currently, she is pursuing the bachelor's degree with the Research Group for Microscopic Image and Medical Image Analysis. Her research interests are in pathology image analysis, random field models and deep learning.

\vspace{1ex}\noindent\textbf{Xiaoyan Li} received her Ph.D. degree in Pathology from China Medical University, China, in 2014. From 2014 till now, she works in the Department of Pathology, Cancer Hospital of China Medical University, Liaoning Cancer Hospital and Institute, engaged in the diagnosis of surgical tumor pathology and molecular pathology, and her main research direction is the occurrence and development mechanism of breast cancer and cervical cancer.

\vspace{1ex}\noindent\textbf{Tao Jiang} received his Ph.D. degree from the University of Siegen, Germany, in 2013. He is currently a Full Professor with the Chengdu University of Information Technology (CUIT), China. He is also the Dean with the Control Engineering College of CUIT. His research interests include machine vision, artificial intelligence, robot control, self-driving auto, and membrane computing.

\end{spacing}
\end{document}